\documentclass[english,doc,natbib]{apa6}
\usepackage[english]{babel}
\usepackage[utf8]{inputenc}  
\usepackage[T1]{fontenc}
\usepackage{amsfonts}
\usepackage{url}   
\usepackage{nameref}
\usepackage{amsmath}  
\usepackage{booktabs}      
\usepackage{tabularx}
\usepackage[colorinlistoftodos, textwidth=40mm, textsize=scriptsize]{todonotes}
\usepackage{verbatim}
\usepackage{graphicx} 
\usepackage{multirow}
\makeatletter

\newcommand*{\inputGraphics}[1]{%
  \ifx\input@path\@undefined
    \def\reserved@a{%
      \let\input@path\Ginput@path
      \InputIfFileExists{#1}{}{file not found}%
      \let\input@path\@undefined
    }%
  \else
    \edef\reserved@a{%
      \noexpand\let\noexpand\input@path\noexpand\Ginput@path
      \noexpand\InputIfFileExists{#1}{}{file not found}%
      \noexpand\def\noexpand\input@path\expandafter{\input@path}%
    }%
  \fi
  \reserved@a
}
\makeatother
\usepackage{floatrow}
\usepackage{subfig}
\usepackage{caption}
\usepackage{multicol}
\usepackage{etoolbox}
\patchcmd{\appendix}{\appendixname}{{\Large Appendix}}{}{}   
\usepackage{placeins}
\usepackage{setspace}% http://ctan.org/pkg/setspace
\AtBeginEnvironment{tabularx}{\singlespacing}
\usepackage{color}

\definecolor{Green}{RGB}{100,200,100}
\definecolor{Blue}{RGB}{0,0,255}
\definecolor{Red}{RGB}{255,0,0}
\definecolor{Orange}{RGB}{255,128,0}

\title{\textbf{Analyzing machine-learned representations: A natural language case study}}
\shorttitle{Analyzing representations}

\fourauthors{Ishita Dasgupta}{Demi Guo}{Samuel J. Gershman}{Noah D. Goodman}
\fouraffiliations{Harvard University}{Harvard University}{Harvard University}{Stanford University}

\authornote{Correspondence concerning this article should be addressed to Ishita Dasgupta,
Department of Physics and Center for Brain Science, Harvard University, 52 Oxford Street, Cambridge, MA 02138, United States. 
E-mail: ishitadasgupta@g.harvard.edu.

A preliminary version of this work was previously reported in \cite{dasgupta2018evaluating}}

\begin{document}

\maketitle

\begin{abstract}

As modern deep networks become more complex, and get closer to human-like capabilities in certain domains, the question arises of how the representations and decision rules they learn compare to the ones in humans. In this work, we study representations of sentences in one such artificial system for natural language processing. We first present a diagnostic test dataset to examine the degree of abstract composable structure represented. Analyzing performance on these diagnostic tests indicates a lack of systematicity in the representations and decision rules, and reveals a set of heuristic strategies. We then investigate the effect of the training distribution on learning these heuristic strategies, and study changes in these representations with various augmentations to the training set. Our results reveal parallels to the analogous representations in people. We find that these systems can learn abstract rules and generalize them to new contexts under certain circumstances -- similar to human zero-shot reasoning. However, we also note some shortcomings in this generalization behavior -- similar to human judgment errors like belief bias. Studying these parallels suggests new ways to understand psychological phenomena in humans as well as informs best strategies for building artificial intelligence with human-like language understanding.

\end{abstract}
\newpage

\section{Introduction}

Recent years have seen a vast improvement in the capabilities of artificial intelligence systems, driven primarily by developments in deep neural networks \citep[see][for a review]{lecun2015deep}. These have allowed artifical system to reach human-level performance at video games \citep{mnih2015human}, object recognition \citep{russakovsky2015imagenet}, and voice generation \citep{oord2016wavenet}, as well as produced impressive performance in several other domains. However, some serious concerns haunt deep learning approaches and their promise as a general solution to artificial intelligence. Many of these concerns surround the lack of structure in the representations and decision criteria these systems learn \citep{marcus2018deep,lake18}. This problem has been implicated in deep learning's data inefficiency and inability to learn abstract structure from few examples, its difficulty in utilizing hierarchical structure and fostering transfer between tasks and domains, as well as the challenge of integrating established prior information into deep learning systems. It also presents serious concerns about the interpretability of its representations and decision criteria, making them less dependable and risky for deployment in sensitive or highly variable domains. 

All of this points to a crucial problem: how can we better understand the representations learned by these systems? Existing studies \citep[e.g.,][]{karpathy2015visualizing, li2015visualizing, yosinski2015understanding, zeiler2014visualizing} primarily use approaches inspired by neuroscience methods developed to understand the brain, for example the statistical analysis of unit activations, and ablation studies where specific units are disconnected or deactivated. These methods promise interesting bottom-up insights into the inner workings of these systems. Cognitive science provides another set of tools to approach this problem from the top down \citep{ritter17, kadar2017representation, mccoy2019right}, by decomposing cognitive processes into their computational components, building models that incorporate these components, and testing these by making predictions about behavior on carefully selected test problems that distinguish different hypotheses. 

The cognitive science approach has yielded huge benefits in understanding higher-level cognition in humans, a prime example of which is the human ability to learn, understand, and produce language \citep{chomsky2002syntactic, linzen2019can}. This domain exemplifies a hallmark of human intelligence: the ability, in the words of von Humboldt, to ``make infinite use of finite means.'' Specifically, human cognitive abilities have been characterized as \textit{systematic} \citep{fodor88, lake2019human} -- this indicates an algebraic capacity to produce new combinations from known components. For example, when a person learns a word in a specific context as part of a particular sentence, they can immediately use this new word in an infinity of other sentences in which this word has never previously been encountered. Systematicity therefore allows humans an impressive capacity to \textit{generalize}, transferring knowledge from one context to others. This ability requires the representations underlying this newly learned word for example, to be abstract (not tied to specific contexts) and compositional (possible to combine with other words and sentences). The failure of neural network models to achieve such systematicity has been a recurring (and controversial) theme in cognitive science \citep{fodor88,lake18}. This concern has previously been studied specifically in the domain of natural language \citep{lake17, gershman15, belinkov2019analysis}, demonstrating the lack of abstract compositional reasoning in certain networks. These analyses are often carried out on toy systems, and while they demonstrate conclusively the lack of systematicity, they largely neglect a deeper analysis of what the systems do learn. 

% redundant with text below
%We advance this line of research by investigating the representations the systems acquire, and the effect of the training distribution on these representations. We also develop novel analyses of systematicity that shed light on the ways in which these machine-learned representations are similar to and different from human representations.

In this paper, we carry out an analysis of the representations learned by a state-of-the-art model for a difficult natural language processing task. We discover that its representations are not systematic; instead, the model uses various heuristic strategies. We then investigate how these heuristics might arise. Analyses of the training distribution reveal that it is very biased, containing many unintended structural regularities that can be exploited by these much simpler heuristics. These simple rules are therefore easily acquired by the neural network, since they explain a substantial amount of variance without having to invoke a more complex systematic representations. We then carry out various augmentations to the training set to better understand if the system can learn abstract composable representations, given the right training distribution. We find parallels between our findings and studies of human representations in terms of how systematic they are under certain circumstances, as well as in terms of when and where this systematicity breaks down. We discuss how such analyses can be fruitful to both cognitive science and machine learning.

% long range syntactic representation to other people studying deep reps -- Tal Linzen etc.

\section{Background}

In this section we review some background on the kinds of representations we will be studying (vector space embeddings of sentences). We also review the three key factors in how such embeddings are generated: the task that they are optimized for, the architecture of the model used to perform that task, and the training distribution on which performance is optimized.\footnote{The details and implementation of the optimization algorithm also contribute \citep[see][for an overview]{ruder2016overview}, but as long as the optimization reaches convergence this has relatively little effect, and we leave this out of our current discussion.}

\subsection{Vector space embeddings}
Vector space models represent items as vectors in some metric space. These have a long history in cognitive science as models of semantic representations \citep{steyvers2006multidimensional, beals1968foundations, pereira2016comparative}. In particular, in the domain of language, vector space models of words (also known as word embeddings) that are learned using distributional information (statistics of text corpora) have been shown to encode syntactic as well as semantic structure, and have been used in psychological models for syntactic category acquisition \citep{redington1998distributional}, inductive vocabulary learning \citep{landauer1997solution}, analogical reasoning \citep{rumelhart1973model}, categorization \citep{jones2007representing}, and high-level associative judgments \citep{bhatia2017associative}. Modern machine learning has allowed the mining of very large datasets to produce vector space embeddings that are now commonly used as the word representations in artificial intelligence systems for natural language processing \citep{pennington14, mikolov2013distributed}.

Understanding language requires understanding not only words, but also their relations within a sentence. These relations are abstract and composable, allowing language to be combinatorially productive -- with a finite set of words, one can systematically produce an infinite set of sentences simply by creating new and longer combinations of these known words. The number of sentences in a language therefore far exceed the number of words. For this reason, generating similar vector embeddings for sentences has proven challenging. Recent papers have developed several supervised as well as unsupervised approaches to learning vector space representations of sentences using recurrent neural networks (RNNs) that are able to represent the order of words in a sentence \citep{kiros2015skip, Hill:2016uu, Conneau:2017uf}. These are intended to capture sentence-level semantic content, and have been shown to perform reasonably well on transfer tasks (sentence-level semantic tasks on which the embeddings were not specifically trained). In particular, the performance of these sentence models exceeds the performance of representations that treat sentences as bags of words (BOW models) -- these patently lack any order information about the words, therefore ignoring the abstract and composable relational structure at the sentence level. However, it is unclear exactly what relational information between words is actually represented in such RNN sentence models. In this work, we start to shed light on this question.

\subsection{Natural Language Inference}
The sentence embeddings we analyze are trained on the natural language inference (NLI) task. The goal is to classify pairs of sentences (a premise and a hypothesis) into `entailment', `contradiction', or `neutral', depending on the semantic relation between the two sentences. This is a popular domain for studying artificial representations since it has a lot of relatively interpretable underlying structure \citep{glockner2018breaking, mccoy2019right, nie2019analyzing}. For example, it is a simple domain in which abstract and composable relational structure is required -- word-level information is not generally sufficient to perform well on this task. The premise sentence ``Anne is more cheerful than Bob'' contradicts the hypothesis sentence ``Anne is less cheerful than Bob'', but entails the hypothesis sentence ``Bob is less cheerful than Anne''. Here, both the hypothesis sentences have the exact same words, and would be indistinguishable if we were just comparing the words in them. More generally, X is more Y than Z entails that Z is less Y than X, for any X, Y and Z. In this case, the specific words used almost don't even matter, and the bulk of the information is in the relations between the words in the sentence. Encoding abstract rules like this allows us to systematically carry out natural language inference on combinatorially many different sentences, with different Xs, Ys, and Zs.

The human ability to carry out abstract reasoning of this sort is a richly studied topic. Some of these abilities however are so obvious, that they are often simply taken for granted without formal study. For example, it is reasonable to assume that any adult human (in the absence of time pressure or cognitive load) can fairly easily process that if X is more Y than Z, then in general Z is less Y than X irrespective of the specific meanings of X, Y and Z. In this paper, we investigate to what extent certain machine-learned sentence embeddings can represent and use such abstract rules in natural language inference.

Despite the generally acknowledged power of human abstract reasoning, a number of studies indicate that humans are not perfect: semantic content (for example the specific meanings of the X, Y and Zs above) has been shown to interfere with systematic inferences in an effect often termed `belief bias' \citep{braine1978relation,johnson1978psychology}. This effect is especially noticeable in children \citep{evans1995belief}, as well as adults under time pressure or cognitive load \citep{evans2013psychology}. In the last part of this paper, we discuss similarities between humans and machines in how they fail certain tests of systematicity.

\begin{figure}
\centering
\includegraphics[width=0.6\textwidth]{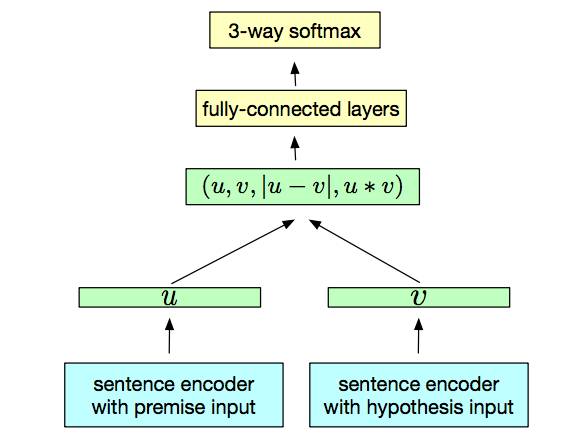}
\caption{InferSent architecture \citep{Conneau:2017uf}.}
\label{fig:arch}
\end{figure}

\subsection{Models for sentence embeddings}
The sentence embeddings we study in this paper are from a highly successful NLI system, InferSent \citep{Conneau:2017uf}. Each premise and hypothesis sentence are input to a sentence encoder as a sequence of pre-trained 300-dimensional GloVe word embeddings \citep{pennington14}. These word embeddings already contain a lot of information about the semantic and syntactic roles of the words (see section on Vector space embeddings for details), and therefore a large part of the lexical information is already represented. Therefore the bulk of the work InferSent has to do is to learn and represent how these words relate to one another in a sentence to provide meanings. The sentence encoder takes in this variable length input and, after passing it through various recurrent and convolutional layers \cite[see][for details]{Conneau:2017uf}, provides a 4096-dimensional vector as output. This output vector serves as a sentence embedding. To make the final inference, these sentence embeddings for the premise and hypothesis are fed to a simple classifier described in Figure \ref{fig:arch} that labels each pair as entailment, neutral or contradiction. The network is trained end-to-end with supervised learning, using a large labelled dataset for NLI (see next section for details on this dataset). The learned embeddings were shown to perform well on other sentence-level tasks (such as sentiment analysis, semantic textual similarity and other natural language inference datasets) by re-using the sentence encoder and training only the classifier for the specific task at hand. This indicates that the sentence encoder does capture some semantic content in the embeddings.

For our tasks, we replicate the procedure in \citet{Conneau:2017uf} to obtain sentence embeddings. These are henceforth referred to as the InferSent sentence embeddings. Our trained InferSent model gives us 84.73\% accuracy on validation and 84.84\% accuracy on the test dataset, which is comparable to the performance of the classifier reported in \citet{Conneau:2017uf}. For comparison, we also train a bag-of-words (BOW) baseline model that averages the pre-trained GloVe word embeddings for all the words in the sentence to form a sentence embedding. These embeddings cannot represent abstract relational structure, since the architecture of the model used to generate them (a simple average of the word embeddings) cannot express word order. We then train a simply classifier on these embeddings to perform natural language inference. This model achieves $53.99\%$ accuracy on the SNLI test set \citep[comparable to the BOW performance reported in][]{Conneau:2017uf}.

Neural networks can act as universal function approximators \citep{siegelmann1995computational, hornik1991approximation}, and given sufficient capacity, they can represent any arbitrarily complex set of relations between the words in the sentence. The InferSent model has a very large capacity due to a large number of layers and hidden units \citep[see][]{Conneau:2017uf}, so a lot of abstract compositional structure is in theory within the representational capacity of these sentence embeddings. In this paper, we analyze how much systematic structure is actually learned and utilized for the NLI task at hand.

% currently a bit degenerate with the abstract - maybe not req?
%\section{Contributions}
%We address the question of what is actually being learned by these sentence embeddings, by designing a diagnostic test dataset (based on comparisons) that relies explicitly on compositional information, and is intractable for systems that capture only lexical information. We characterize the performance of InferSent on this dataset. Given the exponentially large number of possible sentences, most subsets of sentences unless meticulously designed, can be justified by heuristics simpler than the true generative process of language. Performance of InferSent on the diagnostic dataset reveals that InferSent does seem to use some such heuristics when doing Natural Language Inference (NLI). This highlights the value of such structured test datasets in investigating and understanding what exactly some of these complex models encode. We also investigate the ecological validity of these heuristics in the SNLI dataset it is trained on. Finally, we retrain a model on a dataset containing both our comparison-based dataset and the original SNLI train set to see if structured datasets like ours can be learned by the InferSent architecture. This helps determine if the shortfall in compositionality found in InferSent is primarily due to poverty in the training data, or in the model architecture itself and also helps determine future utility of such structured datasets in improving training for models.

\subsection{Training datasets}
To understand sentence embeddings like the ones learned by InferSent, it is imperative to not only consider the model specifications for the system that produces them (in this case the specific end-to-end architecture of the network in InferSent), but also the learning signals it receives from the training set. For many deep learning based methods, very little information about the structure of the task is baked into the architecture of the models -- the only structure about language that it is endowed with before training are the biases that come with using a recurrent neural network as the architecture. This specifies that sentences have variable-length, sequential structure. These embedding models are therefore fairly `tabula rasa', and most of what they represent about the structure of the task (in this case natural language inference) is learned from training data. As elaborated in the previous section, some abstract compositional structure is within the representational capacity of the InferSent sentence embeddings -- but whether or not the right structure is actually learned and represented depends largely on the training data. The significance of the training set on the representations learned by flexible deep learning methods is often not adequately considered. One contribution of this work is to highlight and analyze this issue.
 
InferSent was trained on the Stanford Natural Language Inference (SNLI) dataset \citep{snli:emnlp2015}, a popular labelled dataset for natual language inference. SNLI consists of ~550k premise-hypothesis sentence pairs, and is balanced (consists of equal number of pairs with entailment, contradiction and neutral relationships). The dataset was generated with a crowd-sourcing framework. Workers were presented with a scene description from a corpus of image captions that act as the premise, and asked to supply hypothesis sentences that have each of the three possible NLI relations (entailment, neutral, and contradiction) to the given premise. The freedom to produce entirely novel hypotheses leads to a rich set of sentences; however, it also leads to some artifacts that can strongly bias the representations learned by a `tabula rasa' system. We discuss these in later sections.

\section{A test dataset of minimal cases: The Comparisons dataset}
Our goal is to understand the representations and decision criteria learned by InferSent, in particular how much systematic relational information they encode and utilize -- do they represent abstract rules for the ways words combine to give meaning to sentences? In the machine learning literature on natural language processing, any performance above the bag-of-words (BOW) baseline (that only receives the words in the sentence with no order information) is often seen as proof of the encoding and utilization of relational information. However, this is an unwarranted conclusion---the BOW baseline usually receives only averaged word vectors for the sentence, and therefore also loses some of the lexical information. It often does not actually reach the best possible performance with only the words. Performance above this baseline therefore does not license the conclusion that relational information is being encoded and used at all.

Here, we pursue an alternative approach, inspired by traditions in cognitive psychology and psycholinguistics of building diagnotic test sets to investigate the underlying representations and decision rules. The goal is to generate a set of sentence pairs such that encoding the relations between words (in addition to the words themselves) is \textit{required} to correctly classify them into the three NLI classes. Diagnostic test datasets such as these, that posit a hard baseline for performance without relational information, provide a more foolproof way to test whether such information is being used.

We considered pairs of sentences such that the NLI relation between the sentences can be changed without changing any of the words in the sentence, only their order. We generated our test dataset using comparisons as these are easy to fit into the NLI framework, and yield many simple examples of sentence pairs that require more than word-level data to understand. For example, the premise sentence ``the woman is more cheerful than the man'' contradicts one hypothesis sentence, ``the woman is less cheerful than the man'', but entails another hypothesis sentence, ``the man is less cheerful than the woman''. Since both hypothesis sentences have the exact same words, they would be indistinguishable if we were just comparing their bag-of-words representations. Therefore, a model based only on the words, and not considering the relations between them, would at most get one of the two classifications right. This caps the bag-of-words performance at $50\%$, and some relational rules must be learned to perform above this baseline. 

Generation of several such sentence pairs can be easily automated. We considered three sub-types, described below and summarized in Tables \ref{tab:rules} and \ref{tab:dataset}.
 
\subsection{Same type}
Premise-Hypothesis pairs differ only in the order of the words. \\
{\tt Premise: The woman is more cheerful than the man \\ Hypothesis: The man is more cheerful than the woman \\ CONTRADICTION \\} {\tt Premise: The woman is more cheerful than the man \\ Hypothesis: The woman is more cheerful than the man\\ ENTAILMENT}

\subsection{More-Less type}
Premise-Hypothesis pairs differ by whether they contain the words `more' or `less'. \\
{\tt Premise: The woman is more cheerful than the man \\ Hypothesis: The woman is less cheerful than the man\\ CONTRADICTION \\}{\tt Premise: The woman is more cheerful than the man \\ Hypothesis: The man is less cheerful than the woman \\ ENTAILMENT} 

\subsection{Not type}
Premise-Hypothesis pairs differ by whether they contain the word `not'. \\
{\tt Premise: The woman is more cheerful than the man \\ Hypothesis: The woman is not more cheerful than the man\\ CONTRADICTION \\}{\tt Premise: The woman is more cheerful than the man \\ Hypothesis: The man is not more cheerful than the woman \\ ENTAILMENT}

\begin{table}[htb]
  \begin{center}
%   \resizebox{0.6\columnwidth}{!}{%
  \begin{tabular}{p{20mm} p{45mm} p{45mm}} 
 \toprule
 Type & Entailment hypothesis & Contradiction hypothesis\\
 \midrule
    
Same & 
X is more Y than Z  &
 Z is more Y than X \\
 
More-Less &
Z is less Y than X  &
X is less Y than Z  \\
 
Not &
Z is not more Y than X  &
X is not more Y than Z  \\
\bottomrule
\end{tabular}
% }
 \caption{Rules in Comparisons dataset for Premise: X is more Y than Z}
  \label{tab:rules}
  \end{center}
\end{table}

\begin{table}[htb]
  \begin{center}
  \begin{tabular}{p{60mm} c} 
 \toprule
 Type & Number of sentence pairs \\ [0.5ex] 
 \midrule
  Comparisons (same)  &  14670\\
 Comparisons (more-less)  &  14670\\
 Comparisons (not)  &  14670\\
 \bottomrule
\end{tabular}
 \caption{Comparisons dataset summary.}
  \label{tab:dataset}
  \end{center}
\end{table}

To facilitate comparison with the SNLI dataset, we ensured that the vocabulary distribution of our Comparisons dataset is similar to the original SNLI training dataset.\footnote{Only a few words differed by more than $1 \%$ from their occurrence rate in SNLI, such as \textit{not, a, than, the, is, less, more}. This was inevitable given the general structure of the comparison sentence pairs we use. All of these words however did still occur in the SNLI training corpus, and were not new to the model at test time.} This ensured that we are only manipulating the relational structure of the test set, and poor performance cannot be attributed to not having experienced the specific words before.
% This is to better target the question of whether relational information is being encoded and to ensure that poor performance does not arise from the model simply having not encountered some of the words in our dataset. 
% We do this by controlling the frequency of the words we use to substitute the mouns X and Z and the adjective Y in Table \ref{tab:rules}. 

\section{Testing the sentence embeddings}

We tested the two classifiers based on two different sentence embeddings (the InferSent sentence embeddings, and the BOW sentence embeddings) on the constructed test set (the Comparisons dataset, Table \ref{tab:dataset}). Both of these classifiers were trained for the same task (Natural Language Inference), on the same training dataset (SNLI), and differed only in the model used to generate them. The InferSent embeddings had access to word order, while the BOW embeddings did not (see Section `Models for sentence embeddings' for details). The overall performance of each of the two classifiers on the Comparisons dataset are given in Table \ref{tab:pretrainperf}, and analyzed in greater detail in the following sections.

\begin{table}[htb]
\begin{center}
  \begin{tabular}{p{20mm} p{20mm} p{20mm}} 
 \toprule
 Type & BOW & InferSent \\
    \midrule
same & 50.0 &50.37\\
 more/less & 30.24 &50.35\\
not & 48.98 & 45.24\\
\bottomrule
\end{tabular}
 \caption{Performance on the Comparisons dataset.}
  \label{tab:pretrainperf}
  \end{center}
\end{table}

\subsection{Performance of Bag of Words}
\begin{figure}[ht!]
\centering
\includegraphics[width=0.30\textwidth]{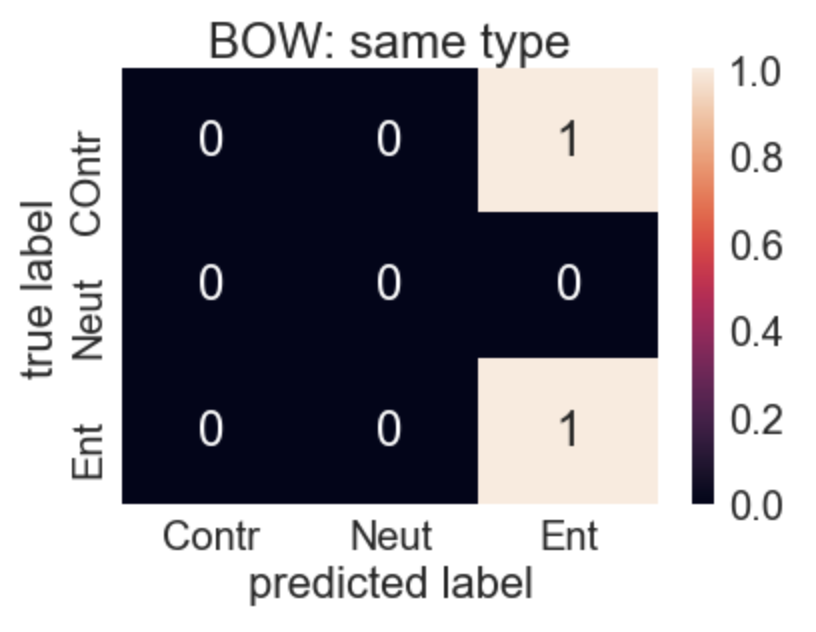}
\includegraphics[width=0.30\textwidth]{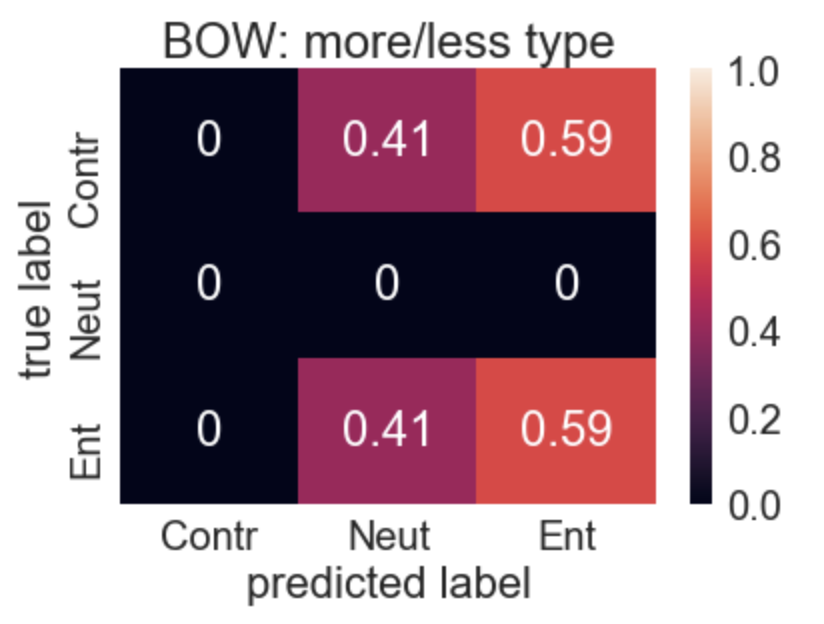}
\includegraphics[width=0.30\textwidth]{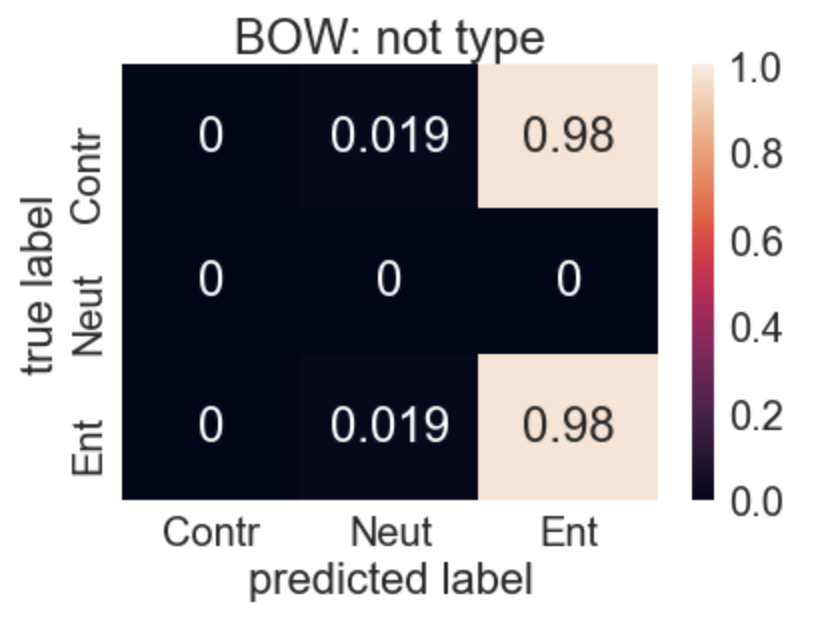}

\caption{BOW embedding confusion matrices, with normalized rows.}

\label{fig:BOWhist}
\end{figure}

We found that the BOW embeddings make classifications that are exactly symmetric across the two true labels (entailment and contradiction) in each task (rows in Figure \ref{fig:BOWhist}). This is expected since the sentence pairs with one label are just permuted versions of the sentence pairs with the other label. Therefore BOW cannot distinguish them, and necessarily classifies both of them the same way. This also ensures that the performance is capped at $50\%$. Asymmetry between the classifications of the two categories can occur only when relational information is encoded in the sentence embedding.

Considering the aggregate performance of BOW in Table \ref{tab:pretrainperf}, we found that performance, particularly on the `more/less' type subset of the test dataset ($30.24\%$), was significantly below $50\%$. This highlights the trouble with using BOW embeddings as a baseline for the encoding and use of relational information. Up to $50\%$ performance is achievable on this dataset without using any relational information; therefore performance above the BOW baseline of $30.24\%$ does not necessarily imply the use of relational information.

\begin{figure}[ht!]
\centering
\includegraphics[width=0.30\textwidth]{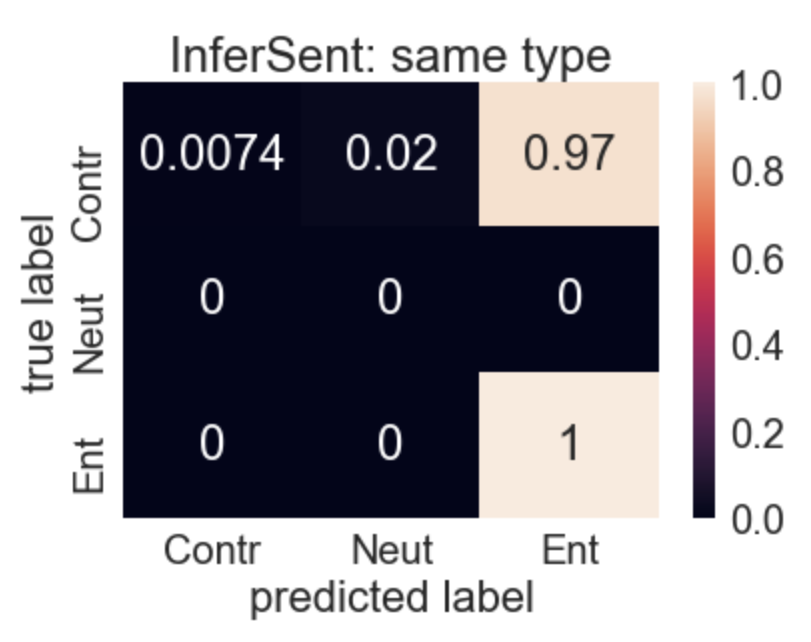}
\includegraphics[width=0.30\textwidth]{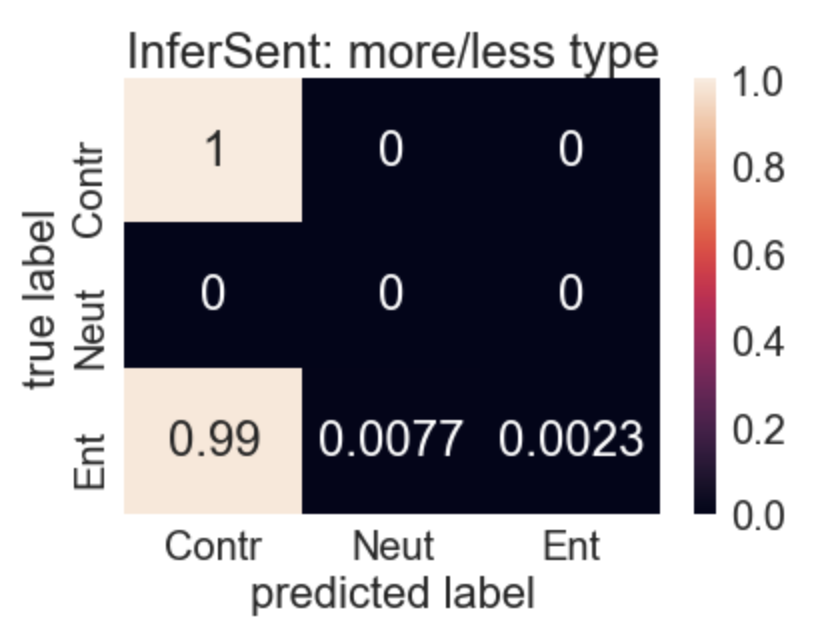}
\includegraphics[width=0.30\textwidth]{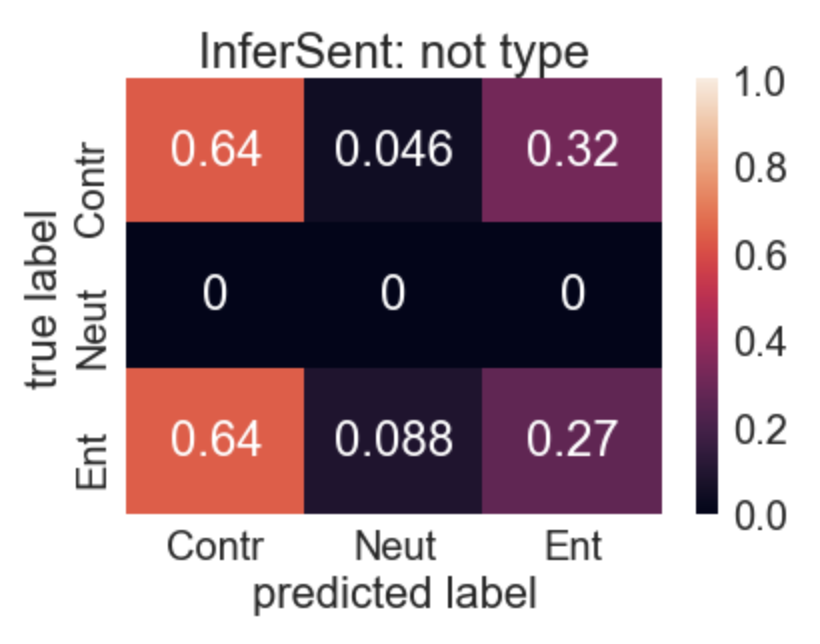}

\caption{InferSent embedding confusion matrices, with normalized rows.}
\label{fig:InferSenthist}
\end{figure}

\subsection{Performance of InferSent}
The performance of the InferSent embeddings was slightly asymmetric (Figure \ref{fig:InferSenthist}), indicating that it was able to distinguish sentences slightly, based on relational information. Yet overall the InferSent embeddings were extremely poor at this task (Table \ref{tab:pretrainperf}), achieving performances slightly above $50\%$ for two of the three sub-types of sentence pairs in the Comparisons dataset, and even less than $50\%$ in a third sub-type. This indicates that InferSent embeddings do not correctly encode and utilize the kinds of abstract relational rules we tested with the Comparisons dataset. 

However, InferSent's performance on another test dataset (the SNLI test dataset) is as high as $84\%$ -- so it is clearly encoding some relevant information about natural language inference. Further, a quick glance at Figure \ref{fig:InferSenthist} indicates that InferSent does not respond randomly to the queries in our Comparisons dataset, but rather in some structured (though incorrect) way. Rather than simply conclude that InferSent embeddings are not systematic and leaving things at that, we can study patterns in the incorrect classifications made to better understand the underlying representations and decisions rules. Since our test dataset is highly structured, it allows a controlled way to generate and test hypotheses about the heuristic representations and decision rules InferSent implements. 

Apart from isolating and characterizing these heuristics, it is also instructive to consider how InferSent might come to encode them in the first place. To answer this, we look to the study of heuristic strategies in humans. The theory of ecological rationality \citep{todd2007environments, simon1991bounded} posits that a system can exploit structural regularities in its learning environment using heuristics that achieve close to optimal performance in that specific environment. These might be much simpler than the most general strategy that performs well in all environments. Heuristics that leverage these structural regularities are therefore termed `ecologically valid' in that environment. This suggests that we can better understand how heuristic strategies might arise in InferSent by examining if they are ecologically valid in its `learning environment' (i.e., the training set). In the following sections, we delve into the heuristic strategies that explain performance on our Comparisons datset, as well as how InferSent might have come to encode them by testing their ecological validity in the SNLI training dataset.

\subsubsection{Overlap Heuristic}
We note in Figure \ref{fig:InferSenthist} that almost all the sentence pairs in the same-type comparisons were classified as entailments, despite half of them being true contradictions. A distinguishing feature of the same-type comparisons is that the premise and hypothesis sentences have full word overlap (they both contain exactly the same words). This observation allows us to hypothesize an \textit{overlap heuristic}: high overlap in words between premise and hypothesis biases InferSent against classifying the pair as a contradiction.

While we have seen some evidence that this heuristic is indeed at play (based on the performance on the same-type comparisons), the question remains as to why it encodes this rule. With our knowledge of language, we know this simple rule to reflect on incorrect understanding of natural language inference. However, all the knowledge about the NLI task that InferSent encodes is from its training dataset. If the dataset has underlying structural regularities that can be exploited by simple heuristic strategies, then a tabula rasa model for NLI such as InferSent that is trained on this dataset will learn to encode it.

We carried out an analysis of the SNLI dataset to determine if the overlap heuristic is ecologically valid in it. First, we observed anecdotally that indeed several contradictory sentence pairs have no overlap in words. For example, a contradictory sentence pair in SNLI is:\\
{\tt Premise: Several people are trying to climb a ladder in a tree. \\ Hypothesis: People are watching a ball game. \\ CONTRADICTION\\}
To quantitatively verify this observation, we ranked all the sentence pairs in SNLI by overlap rate: $\frac{\text{\# of overlap words}}{\text{total \# of words}}$(in non-increasing order). We then considered the top X sentences with highest overlap for different Xs. As shown in Table \ref{tab:highOverlap}, when considering the full dataset, the distribution is balanced (the percentage of entailments, contradictions and neutral sentences are equal). However, we found that as the word overlap in the sentences increases, the percentage of contradictions drops. When considering only the top 1000 sentence pairs for overlap, we found that $91.5\%$ of them have entailment or neutral labels, with only the remaining $8.5\%$ having a contradiction label. 

\begin{table}[htb]
  \begin{center}
  \begin{tabular}{llll}
  \toprule
     Top &Entailment & Neutral & Contradiction \\
    \midrule
All & 33.4 & 33.3 & 33.3 \\
10000 & 39.5 & 35.7 & 24.8\\
 1000 & 50.8 & 40.7 & 8.5\\
 \bottomrule
  \end{tabular}
 \caption{Percentage of entailments split by overlap rate of words in SNLI.}
 \label{tab:highOverlap}
 \end{center}
\end{table}

It is therefore natural that InferSent encodes the simple overlap heuristic as a predictor of contradiction. This explains not only the failure of InferSent to generalize its good performance on SNLI to the same-type comparisons in our test dataset, but also matches the specific failure mode we observe in its responses.

\subsubsection{Antonyms Heuristic}
We note in Figure \ref{fig:InferSenthist} the opposite trend for the more/less-type comparisons, where almost all the sentence pairs were classified as contradictions, despite half of them being true entailments. A distinguishing feature of the more/less-type comparisons is that the the premise and hypothesis always differ by one word -- if the premise contains the word `more' (`less') then the hypothesis always contain the word `less' (`more'). This observation allows us to hypothesize an \emph{antonyms heuristic}: sentences differing in the presence of words that have opposing meanings (antonyms) tend to be classified by InferSent as contradictions, irrespective of the other words or their order in the sentence.

Similarly to the previous section, we investigated the training dataset to elucidate if this heuristic is ecologically valid in InferSent's training set. Anecdotally, we saw that the contradicting hypotheses provided by crowd workers to generate SNLI do follow this pattern. For example, a contradictory sentence pair in SNLI is:\\
{\tt Premise: A man in a white t-shirt takes a picture in the middle of the street with two public buses in the background. \\ Hypothesis:  A man is wearing a black t-shirt. \\ CONTRADICTION\\}

To verify this observation quantitatively, we analyzed the statistics of antonym usage in SNLI. To test whether a sentence pair (A,B) contains antonyms, we went through each word in sentence A, and considered all synonyms of that word, and considered all antonyms of those synonyms. Finally, we checked if sentence B contained any of those antonyms. These synonyms and antonyms were found using the NLTK WordNet software \citep{bird2004nltk}. We then considered two different statistics. First, we calculated P(Contradiction | Antonym), which is the probability that a sentence pair is a contradiction given that its premise and hypothesis contain an antonym pair. This measures how well the presence of antonyms predicts a contradiction label in the training set. Second, we calculated P(Antonym | Contradiction), which is the probability that a contradictory sentence pair contains antonyms. This measures how well a contradiction label predicts antonyms. Both statistics were compared with the equivalent statistic for entailment, to provide a baseline for comparison. Table \ref{tab:antonyms} shows that the presence of antonyms strongly predicts a contradiction label in the SNLI dataset (61.2\% compared to chance at 33.3\%). We also found that a contradiction label predicts the presence of an antonym pair (12.2\%) more strongly than entailment did (3.5\%). This indicates that the antonyms heuristic can explain significant variance for the contradiction label in the training set.

\begin{table}[htb]
  %%\centering
%   \resizebox{0.6\columnwidth}{!}{%
   \begin{tabular}{llll}
  \toprule
    & P(Antonym $|$ X) & P( X $|$ Antonym) \\
    \midrule
    X = Contradiction & 12.2\% & 61.2\% \\
      X = Entailment &  3.5\% &  18.0\% \\
\bottomrule 
  \end{tabular}
%   }
 \caption{ Percentage of entailments split by antonym word pair in the SNLI dataset. 
}
  \label{tab:antonyms}
\end{table}

Since most of our Comparisons dataset contained a large amount of overlap between premise and hypothesis, the rules InferSent applies when responding to these test questions might be biased towards those learned in similar high-overlap settings during training. We checked the statistics of antonymy in the the high overlap subset of SNLI (top 10,000 highest overlap) to provide a closer comparison (Table \ref{tab:antonyms-ho}). Here, contradiction predicts the presence of an antonym pair (43.7\%) more strongly than in the whole dataset (12.2\%). The difference between P(Antonym | Contradiction) and P(Antonym | Entailment) is also more pronounced in this high overlap subset. The presence of an antonym pair no longer predicts contradictions at a high rate (28.9 \%), but this is possibly due to the very low base rate of contradictions in the high overlap subset of SNLI, as compared to entailments.
 
\begin{table}[htb]
  %%\centering
%   \resizebox{0.6\columnwidth}{!}{%
  \begin{tabular}{llll}
  \toprule
    & P(Antonym $|$ X) & P( X $|$ Antonym) \\
    \midrule
    X = Contradiction & 43.5\% & 28.9\% \\
      X = Entailment &  8.7\% &  34.3\% \\
\bottomrule 
  \end{tabular}
%   }
 \caption{ Percentage of entailments split by antonyms in high overlap SNLI subset.}
  \label{tab:antonyms-ho}
\end{table}
 
These results suggest again, that the underlying statistics of the SNLI dataset allow models, including InferSent, to perform well with simple lexical heuristics that ignore the order of words and their relations.

%We also find that the presence of antonyms that are most used in the SNLI dataset are more likely to indicate contradiction for InferSent. For example, the antonyms  women/men, boy/girl and black/white appear more frequently in the SNLI sentence pairs than antonyms like old/young, sitting/standing, and sentences that differ by the first set of antonym pairs are more likely to be classified as contradiction by InferSent, than sentences that differ by the second set of antonym pairs.

\subsubsection{Negation Heuristic}
We see in Figure \ref{fig:InferSenthist} that the not-type comparisons are preferentially classified as contradictions. A distinguishing feature of the not-type comparisons is that the premise and the hypothesis differ by the presence of the negation `not'. This observation allows us to hypothesize a \textit{negation heuristic} where sentence pairs that differ in the presence of negations are preferentially classified as contradictions.

Following procedures analogous to previous sections, we first noted anecdotally, that this heuristic seems to have validity in the contradicting hypotheses in SNLI. For example, a contradictory sentence pair in SNLI is:\\
{\tt Premise: Men turn to the camera to smile on the middle of three long tables in a refectory. \\ Hypothesis:  The man is not smiling. \\ CONTRADICTION\\}

We verified this observation quantitatively by looking at the statistics for negation in SNLI. We collected all sentence pairs that contain ``negating N-grams'': no, not, n't (by considering ``n't'', we included words such as ``don't'' or ``doesn't''). We then carried out analyses similar to the previous section, where we checked (1) the predictive power of negations on contradictions (P(Contradiction | Negation)), and (2) the predictive power of contradiction on negations, P(Negation | Contradiction), and compare both of these to statistics for entailment as a baseline. We found (Table \ref{tab:negation}) that the presence of a negation strongly predicts contradiction in the SNLI dataset (58.4\% compared to chance at 33.3\%). We also found that while both numbers are very low, a contradiction predicts the presence of a negation (3.3\%) slightly more strongly than entailment does (1.1\%). We also carried out the same analysis for a high-overlap subset (top 10,000 highest overlap) of SNLI to maximize similarity with our comparisons dataset and saw similar results (Table \ref{tab:negation-ho}). In fact, the presence of negation predicts a contradiction, P(Negation | Contradiction) = $60.0\%$, at rates comparable to that in the full dataset, P(Negation | Contradiction) = $58.4\%$, despite the much lower base rates of contradiction in this subset of the data. This indicates strong ecological validity for this heuristic in the high overlap subset of the SNLI dataset.

\begin{table}[htb]
  %%\centering
%   \resizebox{0.6\columnwidth}{!}{%
  \begin{tabular}{llll}
  \toprule
    & P(Negation $|$ X) & P( X $|$ Negation) \\
    \midrule
    X = Contradiction & 3.3  & 58.4  \\
      X = Entailment &  1.1  &  20.0  \\
\bottomrule
  \end{tabular}
%   }
 \caption{Percentage of entailments split by negation in SNLI dataset.}
  \label{tab:negation}
\end{table}

\begin{table}[htb]
  %%\centering
%   \resizebox{0.6\columnwidth}{!}{%
  \begin{tabular}{llll}
  \toprule
    & P(Negation $|$ X) & P( X $|$ Negation) \\
    \midrule
    X = Contradiction & 1.3 & 60.0 \\
      X = Entailment &  0.1 &  7.5 \\
\bottomrule 
  \end{tabular}
%   }
 \caption{Percentage of entailments split by negation in high overlap SNLI subset.}
  \label{tab:negation-ho}
\end{table}

%Note that we also did same experiment on more broadly defined negation ngrams which includes Neither, None and etc, and we get similar results. We have $43.6\%$ negation pairs are contradiction, while $27.3\%$ negation pairs are entailment. $7 \%$ contradiction pairs have negation while only $4 \%$ entailment pairs have negation.

\subsubsection{Summary of heuristics}

We found evidence for three heuristics that explain the bulk of the patterns seen in the performance of InferSent on our Comparisons dataset, all of which are ecologically valid in the SNLI dataset. First, we identified the \textit{overlap heuristic} where a large overlap in words between two sentences leads InferSent to not classify them as contradictions. Second, we identified the \textit{antonyms heuristic} and the \textit{negation heuristic}, where the premise and hypothesis differ in the presence of an antonym or a negation, which leads InferSent to classify them as contradictions. 

These illustrate a disproportionate dependence on lexical (rather than relational) meaning in the representations and decision rules used by InferSent. While these heuristics serve well in certain domains, for example in SNLI, they don't amount to a more general encoding of entailment and contradiction between sentence pairs, as evidenced by InferSent's poor performance on our Comparisons dataset.

The analysis so far has highlighted word-level heuristics that InferSent might be using. Yet the confusion matrix results (Figure \ref{fig:InferSenthist}) show a slight asymmetry, indicating at least minor multi-word effects. This suggests that InferSent might be using some (potentially also heuristic) encodings for word order. However, a systematic analysis of the effect of word order, and how much variance such heuristics might explain, is challenging due to the combinatorial explosion in the number of possibilities. We leave a thorough investigation of this to future work.

\section{Augmenting the learning environment}
The foregoing results suggest that such ecological validity of simple heuristics in the SNLI training data (InferSent's learning environment) could explain why InferSent acquires them over a more abstract, systematic representation of the relations between words in a sentence. This leaves open the question of whether architectures such as InferSent are capable of learning the abstract relational rules needed to succeed at our task given a different training set where simple heuristics no longer explain so much of the variance. RNN architectures like the one in InferSent can in theory represent the relational structure required to encode the abstract rules of the sort in Table \ref{tab:rules} (see Section `Models for sentence embeddings' for details). But how might we get them to learn and use them? In this section, we explore this question by training the InferSent model on part of the Comparisons dataset, and testing on a held-out subset of it. This serves to test whether simple training on examples of the rules in Table \ref{tab:rules}, will enable InferSent to encode some abstract relational rules.
% X, Y and Z that it has never previously seen in that combination.

The total training subset of our Comparisons dataset consists of 40k sentence pairs ($7\%$ the size of the 550k pair SNLI training set). Validation and test sets consist of 2000 sentence pairs each. There are no overlapping sentence pairs between any of these sets, therefore simply memorizing the training set will not allow good test performance. Good test performance requires the encoding and utilization of an abstract relational rule.

\begin{table}[htb]
  %%\centering
%   \resizebox{0.6\columnwidth}{!}{%
  \begin{tabular}{llll}
  \toprule
     \multirow{ 2}{*}{Epoch} &
      \multicolumn{3}{c}{Performance (\%)} \\
       & Train(Comp) & Test(Comp) & Test(SNLI)\\
    \midrule
    0 & 47.81 & 45.36 & 84.84\\
    13 & 99.91 & 99.8 & 56.37\\
    \bottomrule
  \end{tabular}
%   }
 \caption{Results of fine-tuning InferSent on the Comparisons dataset.}
  \label{tab:finetuning}
\end{table}

We started with the original InferSent embeddings already trained on the SNLI dataset, and then fine-tuned it on our new Comparisons dataset \citep[using the same protocols used in][to train InferSent]{Conneau:2017uf}. Results are shown in Table \ref{tab:finetuning}. We found that using this method, performance on the SNLI data task degrades over the course of fine-tuning on the new Comparisons dataset from 84.84\% to 56.37\%. This points to over-fitting to the Comparisons data, at the cost of representing information necessary for SNLI. We found however, that performance on the Comparisons test set is much higher (99.8 \%) than when trained only on SNLI (47.81\%). Note that this test set consists of sentence pairs InferSent has never seen before. We thus find that the model architecture for InferSent, given the right training data, can encode some form of abstract relational structure that allows it to learn rules of the form in Table \ref{tab:rules} and apply them to new sentence pairs -- in particular sentence pairs with Xs, Ys and Zs that it has never seen in that combination before.

\begin{table}[htb]
  %%\centering
%   \resizebox{0.6\columnwidth}{!}{%
  \begin{tabular}{llll}
  \toprule
     \multirow{ 2}{*}{Epoch} &
      \multicolumn{3}{c}{Performance (\%)} \\
       & Train(Combined) & Test(Comp) & Test(SNLI)\\
    \midrule
    0 & 33.33 & 33.33 & 33.33\\
    12 & 90.99  & 100.00 & 84.96\\
    %126 & 126 &- 126 -&- 126\\
    %12 & ? & ? & ?\\
    %12 & ? & ? & ?\\
    %12 & 94.12 & 84.34 & 100.0 & 100.0\\
    \bottomrule
  \end{tabular}
%   }
 \caption{Results of retraining Infersent on both SNLI and the Comparisons dataset.}
  \label{tab:retraining}
\end{table}

We then checked whether InferSent can represent this relational structure without losing the information necessary for SNLI. We started with an untrained network, and then trained on an augmented version of the original training data. Here, examples from the SNLI training set were randomly interleaved with examples from our Comparisons training dataset, otherwise using the same training protocols reported in \citep{Conneau:2017uf}. The test results are reported in Table \ref{tab:retraining}. We found that the accuracy obtained this way on the SNLI test set (84.96 \%) is comparable to the model trained only on SNLI (84.84 \%). Moreover, test accuracy on the Comparisons dataset is close to perfect (99.55 \%) and is much higher than the model trained only on SNLI (47.81 \%). This establishes that in this case the model has enough capacity to achieve high performance on specially designed edge-cases like the Comparisons dataset, without loss of performance on the more general SNLI dataset.

This result also verifies that the heuristics we find in the original InferSent are an ecologically rational response to a training environment that licenses these `shortcut' strategies, and not because of shortcomings in representational or learning abilities of the model itself. This points to the benefits of understanding the learning environment in greater detail, and potentially including specially designed data to guard against incorrect heuristics that don't generalize. Research on the generation of adversarial examples targets this intuition. The idea is to have a separate `adversarial' model that generates edge-case training examples optimized to try and fool the main model into giving the wrong answer \citep{goodfellow2014explaining, zhao2017generating}. It does so by generating examples that violate the heuristics the main model has learned from training thus far. Subsequently, the training environment for the model is augmented to include these edge cases making the current heuristics no longer ecologically valid. The main model therefore updates its representations and decision rules accordingly and the process is continued. Our work provides some insight into how we can leverage a top-down understanding of the structure of language and systematic stimulus design, to generate such edge-case training data and potentially improve the representations learned by machine learning systems. 

A key hurdle for the scalability for such augmentation as a solution to improving artificial representations of language however is that there are an infinite number of possible stimuli, with brand new combinations of words that may never have been encountered before. No finite amount of augmentation will allow a system to represent and process this infinite space of natural language sentences unless it can also \textit{generalize} its knowledge gained from the examples observed thus far to new examples. In this section we saw that InferSent can generalize rules like those in Table \ref{tab:rules} to never previously observed combinations of X, Y and Z to perform well on the test set of the Comparisons dataset. In the following sections we further discuss the generalization capacities of the representations learned by InferSent, and focus in particular on their differences and similarities to human generalization.

\section{Generalization}
An important and well-studied aspect of human-like representations is that rules learned with one set of tokens can be systematically generalized to other tokens \citep{fodor88, lake2019human}. In the section on `zero-shot reasoning' we study if our machine-learned representations can perform such generalization to tokens that have never previously been observed. More often however, the tokens to which we want to generalize learned rules have previously been observed, but simply in a different context. The historical contexts of tokens can determine some of their properties -- like syntactic category, and semantic content -- which in turn inform how humans generalize rules to them, sometimes deviating from entirely systematic generalization. In our section on `context-tying', we examine how the historical context of tokens influences systematic generalization in our machine-learned representations, and how these effects compare to those in humans.

% First we examine the role of syntactic category in which tokens restrict generalization to them, and second how the semantic content. 

% For example, many rules of language only generalize within syntactic category -- rules learned about how nouns are used, apply only to other (potentially unobserved and new) nouns, not to other verbs, or adjectives. Further, humans exhibit belief bias across domains \citep{evans1983conflict}, where semantic beliefs about the real-world, can interfere with generalizing abstract logical rules. We study these caveats to human-like generalization, and whether they manifest in our machine-learned representations in a subsection below on context-tying.

Throughout this section, we will only consider sentence pairs that are similar in structure to ones in our Comparisons dataset, and will no longer consider performance on SNLI. We will predominantly be studying the model that has been trained jointly with our Comparisons dataset in addition to SNLI (referred henceforth to as the augmented-InferSent model).

% Thereofore, we only consider the performance of the fine-tuned augmented Infersent model from the previous section, since the retrained augmented model has comparable performance on Comparison-type sentences but is more expensive to train. We refer to it in general as ``the augmented InferSent'' model.

\subsection{Zero-shot reasoning}
Zero-shot reasoning is the ability to solve tasks involving a term that has never been seen before. This (often also called zero-shot learning) has commonly been used as a test for systematicity \citep{lake18} -- a human can carry out inferences like ``Anne is more boffy than Bob'' entails that ``Bob is less boffy than Anne'' without ever having encountered the word ``boffy'' before. 
% Once a person learns a word, they can immediately produce and understand an infinitude of sentences containing it.  It is therefore an extremely efficient way to generalize.

But this ability requires the representation learned to be abstract, and not be tied to the Xs, Ys, and Z's seen in training. Instead it has to encode encode an abstract relational rule where ``X is more Y than Z'' entails ``Z is more Y than X'' for all possible X, Y and Z, irrespective of their specific values. If the representation are tied to the observed values of Xs, Ys and Zs and cannot generalize to new values for these, each possible X, Y and Z has to have occurred in the training dataset. However, these can be arbitrarily complex (e.g., ``The old woman with a flower in her hair \textit{is more deliriously happy than} the tall young man wearing the blue bowler hat'' implies that ``The tall young man wearing the blue bowler hat \textit{is less deliriously happy than} the old woman with a flower in her hair''). Ensuring that every possible such X, Y and Z have been seen in the training data is impossible, and this kind of generalization is key to human-like language understanding.

In this section we consider the performance of the augmented-InferSent model. We already know that this model performs well on both SNLI, and generalizes to new combinations of X, Y, and Z in our Comparisons dataset (see Table \ref{tab:retraining}), where each X, Y and Z have previously been seen. In this section, we analyze its ability to generalize to 3 different kinds of Xs and Zs that have never been encountered during training.

\begin{itemize}
\itemsep0em
    \item \textbf{Held out nouns}: Nouns (from the GloVe dataset) that never occur in the training data (neither SNLI nor our Comparisons dataset).
    \item \textbf{Made up ``words''}: Directly using a 300 dimensional vector randomly sampled from an uncorrelated Gaussian distribution, as a stand-in for a real GloVe vector.
    \item \textbf{Long noun phrases}: The Xs and Zs used in training as part of the Comparisons dataset were of the type ``the man''. Here we generate longer noun phrases of the form ``the grumpy man in front of us'' consisting of randomly sampled adjectives, nouns and prepositional phrases. 
    % Some of these combinations may in theory have been seen in SNLI, but given the length of the noun phrases and the large number of combinations possible, it is very unlikely that the specific phrases used have previously appeared. None of them have been seen in the context of rules like in in the Comparisons dataset (Table \ref{tab:rules}).
\end{itemize}

For each sub-type in the Comparisons dataset (same, more-less and not types), we generated a test set of 1,000 sequences by substituting Xs and Zs of the above kinds. The Ys were sampled in the same way as in the Comparison dataset (random adjectives that appear in SNLI). We then tested on these sentences, and reported the average accuracy. Note that not only had these specific sentences (combinations of X, Y and Z) never been seen during training, even the individual Xs and Zs had not been seen. We found that InferSent generalizes to all three new kinds of Xs and Zs quite well (Table \ref{tab:zsl}). The held-out nouns are the most similar to the Xs and Zs seen during training since they are also exactly one word, and are nouns sampled from GloVe. It is notable that generalization performance with these is comparable to that with the very different kinds of Xs and Zs such as the made-up words, or longer noun phrases, indicating a fairly abstract representation of relational rules that are not tied to the specific value of X and Z.

\begin{table}[h!]
  \begin{center}
    \begin{tabular}{lll}
    \toprule
      Test set  & InferSent ($\%$)&  augmented-InferSent ($\%$)\\
    \midrule
    % Test (Comp) & 100.0 \\
    Held-out nouns & 47.9 & 82.0 \\
    Made up words & 48.0 & 83.2  \\
    Long noun phrases & 49.1  & 84.9 \\
    % Old long noun phrases & 91.1 \\
\bottomrule
\end{tabular}
\end{center}
\caption{Zero-shot reasoning: Performance on previously unobserved Xs and Zs.}
  \label{tab:zsl}
\end{table}

% \ndg{say something about comparison to in-domain generalization (slightly worse?) and comment on the surprising finding that these three types of transfer are about equally well done by the model.} 
This indicates that the representation learned by augmented-InferSent is partially abstract and composable, allowing some systematic generalization to a variety of Xs and Zs that have never been seen before. In the next section we further probe contextuality of generalization and how that interacts with the training set / learning environment, making comparisons to human generalization.

\subsection{Context Tying}
We saw in the previous section that augmented-InferSent has some of the the central human-like capacity of zero-shot reasoning. This indicates some systematicity in its representations. However, even humans do not always succeed at fully systematic generalization. In this section we investigate these exceptions and qualifications to the widest interpretation of systematic generalization, focusing on the role of context in generalization. We do this in two ways: using type violations and biased exposure.

\subsubsection{Type violations}
One extreme of learning a purely abstract rule like in Table \ref{tab:rules} is to be completely insensitive to any properties of the Xs, Ys and Zs, and generalize this rule to all possible tokens. However, this very strong generalization may not always match human intuitions. For example the sentence pair \\
{\tt Premise: The punctual is more cheerful than the man \\ Hypothesis: The punctual is not more cheerful than the man\\}
\noindent does not seem to have a right answer. The rule applies easily only to Xs, Ys and Zs that are of the right type -- in this case the right syntactic category. 

While syntactic structure is not directly provided to the embedding model, some notion of syntactic category will be implicit. Information about the syntactic category of a word can be gleaned from its contexts, i.e. the other words around it \citep{chomsky1993lectures, redington1998distributional, socher2010learning}, and in some cases can be decoded from word embeddings directly \citep{pennington14}.

%We would like to explore the extent to which the system restricts its application of rule patterns based on types. 
%Encoding and utilizing such type information for X, Y and Z when carrying out inferences on them requires that the rules the system learns do not generalize indiscriminately to all Xs, Ys and Zs but rather that to generalize, the test token must satisfy some notion of the right type. 
%Since these systems have no in-built understanding of the roles of different types of tokens, whether a test token is of the `right type' can only be indicated by the contexts in which it has been seen.

We investigated generalization of rules in augmented-Infersent to test items which, unlike in the previous section, had been previously seen, but had only occurred in a different syntactic role (i.e., a different context). We generated a test set of ungrammatical sentences using Xs and Zs that are random non-nouns, in our case random adjectives from SNLI. Crucially, these words had been seen before, but never in the position/context that X and Z occupy in the Comparisons dataset, since appearing in those positions violates syntax. We then evaluated the performance of the augmented-InferSent model in the same way as in the previous section on zero-shot generalization. We found that accuracy on such sentence pairs is low, giving poor performance (Table \ref{tab:type}). This indicates that the rules learned, though at least partially abstract as indicated by generalization to held-out nouns, come with restrictions on the type of (known) items they will apply to. This follows closely how humans generalize -- that learned rules don't generalize indiscriminately to all tokens, but rather only within some fixed categories. These categories in turn, like syntactic categories, can be gleaned from the contexts in which these tokens usually appear. In the next section, we examine the role of semantic content in the context of tokens, and how that influences generalization.

% That this leads to a sensitivity to syntactic type in this case, is reasonable. However, tying tokens to the learning contexts in which they appear in general may not be good. In the next section we study this further.

\begin{table}[h!]
  \begin{center}
    \begin{tabular}{lll}
    \toprule
      Test set  & InferSent ($\%$) & augmented-InferSent ($\%$)\\
    \midrule
    Held-out nouns & 47.9 & 82.0\\
    Non-noun words & 47.9 & 49.3 \\
\bottomrule
\end{tabular}
\end{center}
\caption{Type violations: Performance with tokens from the wrong syntactic category, versus with held-out tokens from the right syntactic category}
  \label{tab:type}
\end{table}

\subsubsection{Biased exposure}

In this section, we manipulate the context of various tokens, without violating the syntactic rules, to study its effect on generalization. In all the augmentations we have used so far, some token X is equally likely to occur in the context of a same-type sentence pair as it is in the context of a more/less-type sentence pair. Similarly, X is as likely to occur in the context where it is `more cheerful than the man' as it is to occur in the context where it is `less cheerful than the man'. Therefore, apart from the restrictions of syntactically correct placement, there is no additional structure around which contexts which tokens occur in -- they are all randomly distributed. 
% However, by virtue of the combinatorially many possible sentences in language, each token is guaranteed to not have occurred in every possible context in which it is syntactically valid. Further, 
However, in the real world, tokens are not uniformly sampled into contexts even within constraints of syntax; a word is much more likely to be sampled repeatedly in certain contexts than others. This is because the appearance of tokens in naturally occurring sentences is not determined solely by their syntactic role, but also by their semantic role.
% The fact that the semantics of sentences are usually used to describe the real world entails that . 
For example, one is more much likely to encounter the sentence ``broccoli is more nutritious than candy'' than the sentence ``candy is more nutritious than broccoli'', since one is true of the real world, and the other is not. Nonetheless, the premise ``candy is more nutritious than broccoli'' still logically entails the hypothesis ``broccoli is less nutritious than candy''. Statistics of how often certain implications and inferences are made in the learning environment (that will be reflected in semantic beliefs about the real-world) can interfere with such logical inferences in humans in both deductive \citep{evans1983conflict} and probabilistic \citep{evans2002background} reasoning. This is often termed `belief bias'. 

% with adults having a reasonably high capacity for abtract inferences even if they violate intutions about the real-world.

In this section, we test if the representations we are studying exhibit belief bias. We manipulate the uniformity in the co-occurrence of tokens with contexts (subject to syntactic constraints), and examine if a newly augmented InferSent model can generalize a token it has seen in one context, to cases where it appears in a different context. We compare this to a zero-shot control condition, where the test token has never been seen before.

To this end, we first generated variants of our Comparisons dataset where tokens are no longer uniformly sampled into contexts. We considered only two sub-types of the comparison types summarized in Table \ref{tab:rules}: the \textit{same-type} ($C_{2}$) and the \textit{more/less-type} ($C_{1}$). These consist the two contexts $C_{1}$ and $C_{2}$ in which tokens can appear. Noun phrases were generated using the same procedure used for the long noun phrases in the section on zero-shot reasoning---phrases (tokens) of the form ``the grumpy man in front of us''. These tokens were then randomly divided into $T^{0}$-type and $T^{*}$-type (460 each). Therefore there is no structural difference between the the $T^{0}$ and $T^{*}$ tokens, only the context in which they are seen will differ across conditions. 
% 500 Glove nouns from spacy... we manually curated a list of ~200 adjectives and ~50 proposition phrases. Then, we divide our noun, adjective and proposition phrases lists into two groups. Finally, we create two groups of noun phrases by randomly concatenating noun, adjective and proposition phrases following some manually curated templates (e.g. first sample a function word such as the, our, her. Then, Sample multiple adjectives. Finally, sample a noun)

We built four sets of sentence pairs that vary in their context-token combination: $C_{2}T^{0}$ consisted of combinations of $T^{0}$ tokens in a $C_{2}$ context, so on and so forth for $C_{2}T^{*}$, $C_{1}T^{0}$, and $C_{1}T^{*}$. Each such context-token combination set was independently divided into train and test sets (each of size 5000). The sentence pairs in each of the four test sets had never been seen before in any of the four training sets. 

We augmented the original InferSent embeddings with different combinations of samples from the four different train sets.\footnote{In this experiment we only make comparisons between the performances of differently augmented models, rather than considering the overall performance like in previous experiments. The influence on performance from the SNLI training data is irrelevant since it will affect all four augmented models equally. Therefore we can neglect SNLI performance and carry out our experiments using fine-tuned augmentation rather than full retraining (see the Section `Augmenting the learning environment' for details on these). This is computationally a lot cheaper.} We then compared their performance on all four of the test sets to examine how different context-token combinations seen during training influenced test generalization. The three different embeddings that result are as follows:
\vspace{-2mm}
\begin{itemize}
\itemsep0em
    \item \textbf{Zero-shot control condition}: Only the $T^{0}$ tokens were seen in training; no $T^{*}$ token were seen at all. Therefore testing with tokens from $T^{*}$ is analogous to zero-shot reasoning. The training set consisted of the full training sets from $C_{1}T^{0}$ and $C_{2}T^{0}$.
    \item \textbf{Experimental conditions}: Both $T^{0}$ and $T^{*}$ tokens were seen in training, therefore testing with tokens from $T^{*}$ is not analogous to zero-shot reasoning. However, the contexts in which $T^{0}$ and $T^{*}$ tokens appear during training differed. There are two different embeddings we trained of this kind.
    \begin{itemize}
        \item \textbf{Exposed--$C_{1}T^{*}$ }: This embedding saw $T^{0}$ tokens in both $C_{1}$ and $C_{2}$ contexts (as with the control condition), and additionally also saw $T^{*}$ tokens -- but only in the $C_{1}$ context.
        In order to balance the number of training examples from each context between conditions, the training set consisted of the full training sets from $C_{2}T^{0}$ and half (randomly selected) of the training set from each of the $C_{1}T^{0}$ and $C_{1}T^{*}$ context-token combination sets.
        \item \textbf{Exposed--$C_{2}T^{*}$ }: This embedding saw $T^{0}$ tokens in both $C_{1}$ and $C_{2}$ contexts, but saw $T^{*}$ tokens only in the $C_{2}$ context. The training set was balanced across contexts here as well.
    \end{itemize}
\end{itemize}

\begin{table}[h!]
  \begin{center}
%   \resizebox{0.6\columnwidth}{!}{%
    \begin{tabular}{llll}
    \toprule
     \multirow{2}{*}{Test set} &
      \multicolumn{3}{c}{Performance (\%)} \\
       & Zero-shot  & Exposed--$C_{1}T^{*}$ & Exposed--$C_{2}T^{*}$ \\
    \midrule
    $C_{1}T^{0} + C_{2}T^{0}$ & 97.44 & 97.02 & 98.0\\
    
    $C_{1}T^{*}$  & 95.72 & 99.7 & \textbf{61.16}  \\
    
     $C_{2}T^{*}$ & 95.78 & \textbf{67.71} & 99.96 \\
    
\bottomrule
\end{tabular}
% }
\end{center}
\caption{Biased exposure: Results from InferSent embeddings augmented with different training sets that manipulate the co-occurrence of context and token.}
  \label{tab:comb}
\end{table}

All three models received the same number of training examples, with equal numbers of sentence pairs from both contexts $C_{1}$ and $C_{2}$. They all also saw $T^{*}$ noun phrases appear in both contexts. The three models only differed in which contexts $T^{*}$ noun phrases appeared during training. The control model never saw $T^{*}$ noun phrases, Exposed--$C_{1}T^{*}$  only saw them in the $C_1$ context and Exposed--$C_{2}T^{*}$ only saw them in the $C_2$ context. All of these were then tested on the same held-out test set. We see from Table \ref{tab:comb} that all three models generalize well to held-out test examples involving previously unobserved combinations of $T^{0}$ noun phrases in both contexts (first row). This is consistent with our initial results on augmentation (see section `Augmenting the learning environment'). Further, the control (zero-shot reasoning) condition that never saw $T^{*}$ noun phrases in training generalizes well to all the test examples with $T^{*}$ noun phrases (first column). This is consistent with our results on zero-shot generalization (see section `Zero-shot reasoning'). 

We now turn to generalization performance when tokens were seen before but only in a specific context (second and third columns in Table \ref{tab:comb}). We discuss the results for the model Exposed--$C_{1}T^{*}$ (that saw $T^{*}$ noun phrases in $C_{1}$ type comparisons), a symmetric discussion applies also to Exposed--$C_{2}T^{*}$.  W found that Exposed--$C_{1}T^{*}$ performs well on held-out test examples from the $C_{1}T^{*}$ category (99.7 \%) -- as consistent with our original experiments with augmentation. However, we found that it fails to generalize very well to $T^{*}$ type noun phrases in the $C_{2}$ context, with a significant drop in performance (67.71 \%). The crucial comparison is that this low performance is also significantly worse that that of the zero-shot control on the same test set (95.78 \%). Neither of these have seen $T^{*}$ phrases in the $C_{2}$ context -- yet the control generalizes very well, while the Exposed--$C_{1}T^{*}$ fails to. This indicates that while the representations learned can generalize well to previously unseen tokens, this generalization is poorer to tokens that have in fact been seen before, but only in a different context.

This indicates that our representations do learn something akin to belief bias, where the context in which tokens have been seen (even within the right syntactic category) can influence how abstract logical rules (like in Table \ref{tab:rules}) generalize to them. This suggests potential directions for research on modeling how belief bias in humans arises. However, it is crucial to point out that although humans do exhibit such context tying, the effects are mostly observed in children \citep{evans1995belief} and under time pressure / cognitive load \citep{evans2005rapid}. The co-existence of such a fast heuristic strategy (that potentially suffers from belief bias), and a slower deliberative strategy (that can perform abstract reasoning) is a well-studied and popular model for representations and decision rules in humans \citep{evans2005rapid, kahneman2011thinking, groves1970habituation}. Thus, although people have a tendency towards belief bias, they are able to overcome it and engage in abstract reasoning, which our machine-learned representations cannot do.

This raises a new concern about the scalability of augmentation as a general approach to learning systematic representations in such tabula rasa machine-learning systems. There are infinitely many possible sentences that all follow the rules of syntax, so observing tokens in contexts that one has not often seen them in, but where they are syntactically valid, is likely to occur often. Our new findings show that while zero-shot reasoning to previously unobserved tokens works in certain cases, these tabula rasa systems may tie an observed token to the small fraction of contexts in which it has been seen. This hinders generalization to cases where this token occurs in a new context. In order for every token to have been observed in every context, a combinatorially large amount of augmented training data would be required, potentially making this approach unfeasible for achieving the kinds of systematic representations humans have.
 
\section{Discussion and Future Work}

In this paper, we carried out a case study in the use of methods from cognitive science and psycholinguistics to better understand machine-learned representations. We developed minimal cases in a natural language inference task that test for some aspects of abstract relational structure in sentences. We used this diagnostic tool on large-scale state-of-the-art sentence embeddings \citep{Conneau:2017uf} to not only demonstrate its lack of abstract composable structure, but also provide insight into the representations and decision criteria actually learned. This approach led us to isolate the use of some simple heuristics, which we then traced to structural regularities in the training distribution. This allowed us to demonstrate the strong effect the training data has on the representations learned. We then augmented this training environment with so-called adversarial examples such that simple heuristics like the ones we found are no longer ecologically valid. We found that such augmentation leads the system to learn some forms of abstract relational structure. Notably, we found that one of the traditional holy grails of systematicity ---zero-shot generalization of learned rules to new, previously unseen words---can be partially  achieved using appropriate augmentation. Further tests, however, revealed limitations to the breadth of this generalization. We found that while zero-shot generalization to previously unseen words works, generalizations to words that have previously been seen in a different context, suffers. This gives us another measure for the extent of systematicity in representations---a phenomena we call `context-tying'. We discussed the relationship between this effect and findings in human cognitive psychology where semantic beliefs about the real-world can interfere with flexible inferences supported by abstract logical representations \citep{evans2013psychology}. This parallel suggests new ways to model this psychological phenomena \citep{dasgupta2019theory}. The presence of context-tying in the machine-learned representations indicates that combinatorially large amounts of augmentation will likely be required for a tabula rasa unstructured neural network model to learn an entirely systematic representation from data. 
% This points to the value of utilizing insights from the study of such representations in humans to build in appropriate inductive biases \citep{lake18}.

These results suggest many directions for future work. We showed how the issue of context-tying bodes poorly for the scalability of using only training set augmentations to achieve human-like systematic representations. Recent work, however, suggests such adversarial mechanisms in the human brain \citep{gershman2019generative}. This motivates further research on how this approach might be made more scalable. We studied the representations learned from a fixed amount of augmentation and training. An important step forward is to better understand how systematicity in these representations evolves over the course of augmented training, and exactly how much augmentation is really needed. Another important problem is to understand what augmentations work best. To that end, a promising direction is to integrate our approach, where augmentations are generated using existing knowledge about analogous representations in humans, with approaches that learn to generate such adversarial augmentations \citep{kang2018adventure,goodfellow2014explaining, zhao2017generating}.

Human infants are not as tabula rasa as models like InferSent but rather encode useful inductive biases \citep{mitchell1980need, pearl2016statistical, chomsky2002syntactic, lightfoot1984language, seidenberg1997language}. Building such biases into our models \citep{lake18, gandhi2019mutual, dubey2018investigating, battaglia2018relational} is a promising direction towards scalably learning systematic representations. We also showed how analysis and controlled testing for heuristic strategies in the learning environment can provide rich insights into the representations learned. Such analyses could also be used to improve learning and subsequent performance by leveraging this underlying structure \citep{csimcsek2013linear, csimcsek2016most, gigerenzer1999simple, martignon2002fast}. Finally, we leverage methods from cognitive psychology to introduce a new structured test dataset (the Comparisons dataset) as well as a new metric (context-tying) for sentence representations. Rather than the traditional single-dimensional metrics of the accuracy achieved on ad-hoc test datasets, our approach provides insights into the kinds of mistakes made and therefore a more principled and nuanced ways to benchmark artificial systems against humans \citep{white2017inference, marelli2014semeval, lake17, linzen2016assessing, mccoy2019right, glockner2018breaking}. A metric like context-tying is not bound to the domain of language, and can also be used to benchmark systematicity in other domains that benefit from abstract compositional representations -- like scene understanding \citep{ommer2009learning, johnson2017clevr} or structured planning \citep{burridge1999sequential, singh1992transfer}. Future work should pursue other such diagnostic metrics, to build towards a comprehensive suite of testable criteria for exactly what constitutes human-like representations, and also to further inform which aspects of these we wish to emulate in artificial systems.

% \ndg{this final paragraph is largey redundant with the above, and kind of a flat ending. maybe replace it with a shorter speculation toward future work and implications?}
% Our main contributions are 1) to show a proof-of-concept of how methods from cognitive psychology can be applied to understanding representations in artificial systems, particularly those trained for high level cognitive functions like language which cognitive scientists have studied for decades, 2) to show how this approach can build mutually-benefical bridges between the two literatures -- for example studies in psychology of ecological rationality and belief bias can inspire new experiments to further analyze the representations learned by these systems, and the fact that these phenomena also show up in fully engineered artificial systems provides inroads to study and model these effects in humans, and 3) to find and introduce tangible tests and measures, like the extent of `context-tying', that allow a more nuanced understanding of what it means to achieve systematicity.

% This indicates that without building in to some extent the kinds of structures we know, from studies in psychology and linguistics, to exist in language, it may be unrealistic for vector space models to learn, with finite data, the abstract representations required for the most general form of human-like language understanding.

\subsection{Acknowledgements}

We are grateful to Anatole Gershman, Tim O'Donnell, Joshua Greene and Fiery Cushman for helpful discussions. ID is supported by Microsoft Research. SJG is supported by the Office of Naval Research (N00014-17-1-2984). NDG is supported by DARPA agreement number FA8750-14-2-0009, and a Sloan Foundation Research Fellowship. 
% Code and data are available at: \url{github.com/ishita-dg/ScrambleTests/tree/training-experiment/testData/new}.

\bibliographystyle{apa-good}
\bibliography{library}

\begin{thebibliography}{}

\bibitem [\protect \citeauthoryear {%
Battaglia%
\ \protect \BOthers {.}}{%
Battaglia%
\ \protect \BOthers {.}}{%
{\protect \APACyear {2018}}%
}]{%
battaglia2018relational}
\APACinsertmetastar {%
battaglia2018relational}%
\begin{APACrefauthors}%
Battaglia, P\BPBI W.%
, Hamrick, J\BPBI B.%
, Bapst, V.%
, Sanchez-Gonzalez, A.%
, Zambaldi, V.%
, Malinowski, M.%
\BDBL {}others%
\end{APACrefauthors}%
\unskip\
\newblock
\APACrefYearMonthDay{2018}{}{}.
\newblock
{\BBOQ}\APACrefatitle {Relational inductive biases, deep learning, and graph
  networks} {Relational inductive biases, deep learning, and graph
  networks}.{\BBCQ}
\newblock
\APACjournalVolNumPages{arXiv preprint arXiv:1806.01261}{}{}{}.
\PrintBackRefs{\CurrentBib}

\bibitem [\protect \citeauthoryear {%
Beals%
, Krantz%
\BCBL {}\ \BBA {} Tversky%
}{%
Beals%
\ \protect \BOthers {.}}{%
{\protect \APACyear {1968}}%
}]{%
beals1968foundations}
\APACinsertmetastar {%
beals1968foundations}%
\begin{APACrefauthors}%
Beals, R.%
, Krantz, D\BPBI H.%
\BCBL {}\ \BBA {} Tversky, A.%
\end{APACrefauthors}%
\unskip\
\newblock
\APACrefYearMonthDay{1968}{}{}.
\newblock
{\BBOQ}\APACrefatitle {Foundations of multidimensional scaling.} {Foundations
  of multidimensional scaling.}{\BBCQ}
\newblock
\APACjournalVolNumPages{Psychological review}{75}{2}{127}.
\PrintBackRefs{\CurrentBib}

\bibitem [\protect \citeauthoryear {%
Belinkov%
\ \BBA {} Glass%
}{%
Belinkov%
\ \BBA {} Glass%
}{%
{\protect \APACyear {2019}}%
}]{%
belinkov2019analysis}
\APACinsertmetastar {%
belinkov2019analysis}%
\begin{APACrefauthors}%
Belinkov, Y.%
\BCBT {}\ \BBA {} Glass, J.%
\end{APACrefauthors}%
\unskip\
\newblock
\APACrefYearMonthDay{2019}{}{}.
\newblock
{\BBOQ}\APACrefatitle {Analysis methods in neural language processing: A
  survey} {Analysis methods in neural language processing: A survey}.{\BBCQ}
\newblock
\APACjournalVolNumPages{Transactions of the Association for Computational
  Linguistics}{7}{}{49--72}.
\PrintBackRefs{\CurrentBib}

\bibitem [\protect \citeauthoryear {%
Bhatia%
}{%
Bhatia%
}{%
{\protect \APACyear {2017}}%
}]{%
bhatia2017associative}
\APACinsertmetastar {%
bhatia2017associative}%
\begin{APACrefauthors}%
Bhatia, S.%
\end{APACrefauthors}%
\unskip\
\newblock
\APACrefYearMonthDay{2017}{}{}.
\newblock
{\BBOQ}\APACrefatitle {Associative judgment and vector space semantics.}
  {Associative judgment and vector space semantics.}{\BBCQ}
\newblock
\APACjournalVolNumPages{Psychological review}{124}{1}{1}.
\PrintBackRefs{\CurrentBib}

\bibitem [\protect \citeauthoryear {%
Bird%
\ \BBA {} Loper%
}{%
Bird%
\ \BBA {} Loper%
}{%
{\protect \APACyear {2004}}%
}]{%
bird2004nltk}
\APACinsertmetastar {%
bird2004nltk}%
\begin{APACrefauthors}%
Bird, S.%
\BCBT {}\ \BBA {} Loper, E.%
\end{APACrefauthors}%
\unskip\
\newblock
\APACrefYearMonthDay{2004}{}{}.
\newblock
{\BBOQ}\APACrefatitle {NLTK: the natural language toolkit} {Nltk: the natural
  language toolkit}.{\BBCQ}
\newblock
\BIn{} \APACrefbtitle {Proceedings of the ACL 2004 on Interactive poster and
  demonstration sessions} {Proceedings of the acl 2004 on interactive poster
  and demonstration sessions}\ (\BPG~31).
\PrintBackRefs{\CurrentBib}

\bibitem [\protect \citeauthoryear {%
Bowman%
, Angeli%
, Potts%
\BCBL {}\ \BBA {} Manning%
}{%
Bowman%
\ \protect \BOthers {.}}{%
{\protect \APACyear {2015}}%
}]{%
snli:emnlp2015}
\APACinsertmetastar {%
snli:emnlp2015}%
\begin{APACrefauthors}%
Bowman, S\BPBI R.%
, Angeli, G.%
, Potts, C.%
\BCBL {}\ \BBA {} Manning, C\BPBI D.%
\end{APACrefauthors}%
\unskip\
\newblock
\APACrefYearMonthDay{2015}{}{}.
\newblock
{\BBOQ}\APACrefatitle {A large annotated corpus for learning natural language
  inference} {A large annotated corpus for learning natural language
  inference}.{\BBCQ}
\newblock
\BIn{} \APACrefbtitle {Proceedings of the 2015 Conference on Empirical Methods
  in Natural Language Processing (EMNLP).} {Proceedings of the 2015 conference
  on empirical methods in natural language processing (emnlp).}
\newblock
\APACaddressPublisher{}{Association for Computational Linguistics}.
\PrintBackRefs{\CurrentBib}

\bibitem [\protect \citeauthoryear {%
Braine%
}{%
Braine%
}{%
{\protect \APACyear {1978}}%
}]{%
braine1978relation}
\APACinsertmetastar {%
braine1978relation}%
\begin{APACrefauthors}%
Braine, M\BPBI D.%
\end{APACrefauthors}%
\unskip\
\newblock
\APACrefYearMonthDay{1978}{}{}.
\newblock
{\BBOQ}\APACrefatitle {On the relation between the natural logic of reasoning
  and standard logic.} {On the relation between the natural logic of reasoning
  and standard logic.}{\BBCQ}
\newblock
\APACjournalVolNumPages{Psychological review}{85}{1}{1}.
\PrintBackRefs{\CurrentBib}

\bibitem [\protect \citeauthoryear {%
Burridge%
, Rizzi%
\BCBL {}\ \BBA {} Koditschek%
}{%
Burridge%
\ \protect \BOthers {.}}{%
{\protect \APACyear {1999}}%
}]{%
burridge1999sequential}
\APACinsertmetastar {%
burridge1999sequential}%
\begin{APACrefauthors}%
Burridge, R\BPBI R.%
, Rizzi, A\BPBI A.%
\BCBL {}\ \BBA {} Koditschek, D\BPBI E.%
\end{APACrefauthors}%
\unskip\
\newblock
\APACrefYearMonthDay{1999}{}{}.
\newblock
{\BBOQ}\APACrefatitle {Sequential composition of dynamically dexterous robot
  behaviors} {Sequential composition of dynamically dexterous robot
  behaviors}.{\BBCQ}
\newblock
\APACjournalVolNumPages{The International Journal of Robotics
  Research}{18}{6}{534--555}.
\PrintBackRefs{\CurrentBib}

\bibitem [\protect \citeauthoryear {%
Chomsky%
}{%
Chomsky%
}{%
{\protect \APACyear {1993}}%
}]{%
chomsky1993lectures}
\APACinsertmetastar {%
chomsky1993lectures}%
\begin{APACrefauthors}%
Chomsky, N.%
\end{APACrefauthors}%
\unskip\
\newblock
\APACrefYear{1993}.
\newblock
\APACrefbtitle {Lectures on government and binding: The Pisa lectures}
  {Lectures on government and binding: The pisa lectures}\ (\BNUM~9).
\newblock
\APACaddressPublisher{}{Walter de Gruyter}.
\PrintBackRefs{\CurrentBib}

\bibitem [\protect \citeauthoryear {%
Chomsky%
\ \BBA {} Lightfoot%
}{%
Chomsky%
\ \BBA {} Lightfoot%
}{%
{\protect \APACyear {2002}}%
}]{%
chomsky2002syntactic}
\APACinsertmetastar {%
chomsky2002syntactic}%
\begin{APACrefauthors}%
Chomsky, N.%
\BCBT {}\ \BBA {} Lightfoot, D\BPBI W.%
\end{APACrefauthors}%
\unskip\
\newblock
\APACrefYear{2002}.
\newblock
\APACrefbtitle {Syntactic structures} {Syntactic structures}.
\newblock
\APACaddressPublisher{}{Walter de Gruyter}.
\PrintBackRefs{\CurrentBib}

\bibitem [\protect \citeauthoryear {%
Conneau%
, Kiela%
, Schwenk%
, Barrault%
\BCBL {}\ \BBA {} Bordes%
}{%
Conneau%
\ \protect \BOthers {.}}{%
{\protect \APACyear {2017}}%
}]{%
Conneau:2017uf}
\APACinsertmetastar {%
Conneau:2017uf}%
\begin{APACrefauthors}%
Conneau, A.%
, Kiela, D.%
, Schwenk, H.%
, Barrault, L.%
\BCBL {}\ \BBA {} Bordes, A.%
\end{APACrefauthors}%
\unskip\
\newblock
\APACrefYearMonthDay{2017}{{\APACmonth{05}}}{}.
\newblock
{\BBOQ}\APACrefatitle {{Supervised Learning of Universal Sentence
  Representations from Natural Language Inference Data}} {{Supervised Learning
  of Universal Sentence Representations from Natural Language Inference
  Data}}.{\BBCQ}
\newblock

\PrintBackRefs{\CurrentBib}

\bibitem [\protect \citeauthoryear {%
Dasgupta%
, Guo%
, Stuhlm{\"u}ller%
, Gershman%
\BCBL {}\ \BBA {} Goodman%
}{%
Dasgupta%
\ \protect \BOthers {.}}{%
{\protect \APACyear {2018}}%
}]{%
dasgupta2018evaluating}
\APACinsertmetastar {%
dasgupta2018evaluating}%
\begin{APACrefauthors}%
Dasgupta, I.%
, Guo, D.%
, Stuhlm{\"u}ller, A.%
, Gershman, S\BPBI J.%
\BCBL {}\ \BBA {} Goodman, N\BPBI D.%
\end{APACrefauthors}%
\unskip\
\newblock
\APACrefYearMonthDay{2018}{}{}.
\newblock
{\BBOQ}\APACrefatitle {Evaluating compositionality in sentence embeddings}
  {Evaluating compositionality in sentence embeddings}.{\BBCQ}
\newblock
\APACjournalVolNumPages{Proceedings of the 39th Annual Conference of the
  Cognitive Science Society}{}{}{}.
\PrintBackRefs{\CurrentBib}

\bibitem [\protect \citeauthoryear {%
Dasgupta%
, Schulz%
, Tenenbaum%
\BCBL {}\ \BBA {} Gershman%
}{%
Dasgupta%
\ \protect \BOthers {.}}{%
{\protect \APACyear {2019}}%
}]{%
dasgupta2019theory}
\APACinsertmetastar {%
dasgupta2019theory}%
\begin{APACrefauthors}%
Dasgupta, I.%
, Schulz, E.%
, Tenenbaum, J\BPBI B.%
\BCBL {}\ \BBA {} Gershman, S\BPBI J.%
\end{APACrefauthors}%
\unskip\
\newblock
\APACrefYearMonthDay{2019}{}{}.
\newblock
{\BBOQ}\APACrefatitle {A theory of learning to infer} {A theory of learning to
  infer}.{\BBCQ}
\newblock
\APACjournalVolNumPages{BioRxiv}{}{}{644534}.
\PrintBackRefs{\CurrentBib}

\bibitem [\protect \citeauthoryear {%
Dubey%
, Agrawal%
, Pathak%
, Griffiths%
\BCBL {}\ \BBA {} Efros%
}{%
Dubey%
\ \protect \BOthers {.}}{%
{\protect \APACyear {2018}}%
}]{%
dubey2018investigating}
\APACinsertmetastar {%
dubey2018investigating}%
\begin{APACrefauthors}%
Dubey, R.%
, Agrawal, P.%
, Pathak, D.%
, Griffiths, T\BPBI L.%
\BCBL {}\ \BBA {} Efros, A\BPBI A.%
\end{APACrefauthors}%
\unskip\
\newblock
\APACrefYearMonthDay{2018}{}{}.
\newblock
{\BBOQ}\APACrefatitle {Investigating human priors for playing video games}
  {Investigating human priors for playing video games}.{\BBCQ}
\newblock
\APACjournalVolNumPages{arXiv preprint arXiv:1802.10217}{}{}{}.
\PrintBackRefs{\CurrentBib}

\bibitem [\protect \citeauthoryear {%
Evans%
}{%
Evans%
}{%
{\protect \APACyear {2013}}%
}]{%
evans2013psychology}
\APACinsertmetastar {%
evans2013psychology}%
\begin{APACrefauthors}%
Evans, J\BPBI S\BPBI B.%
\end{APACrefauthors}%
\unskip\
\newblock
\APACrefYear{2013}.
\newblock
\APACrefbtitle {The psychology of deductive reasoning (Psychology revivals)}
  {The psychology of deductive reasoning (psychology revivals)}.
\newblock
\APACaddressPublisher{}{Psychology Press}.
\PrintBackRefs{\CurrentBib}

\bibitem [\protect \citeauthoryear {%
Evans%
, Barston%
\BCBL {}\ \BBA {} Pollard%
}{%
Evans%
\ \protect \BOthers {.}}{%
{\protect \APACyear {1983}}%
}]{%
evans1983conflict}
\APACinsertmetastar {%
evans1983conflict}%
\begin{APACrefauthors}%
Evans, J\BPBI S\BPBI B.%
, Barston, J\BPBI L.%
\BCBL {}\ \BBA {} Pollard, P.%
\end{APACrefauthors}%
\unskip\
\newblock
\APACrefYearMonthDay{1983}{}{}.
\newblock
{\BBOQ}\APACrefatitle {On the conflict between logic and belief in syllogistic
  reasoning} {On the conflict between logic and belief in syllogistic
  reasoning}.{\BBCQ}
\newblock
\APACjournalVolNumPages{Memory \& cognition}{11}{3}{295--306}.
\PrintBackRefs{\CurrentBib}

\bibitem [\protect \citeauthoryear {%
Evans%
\ \BBA {} Curtis-Holmes%
}{%
Evans%
\ \BBA {} Curtis-Holmes%
}{%
{\protect \APACyear {2005}}%
}]{%
evans2005rapid}
\APACinsertmetastar {%
evans2005rapid}%
\begin{APACrefauthors}%
Evans, J\BPBI S\BPBI B.%
\BCBT {}\ \BBA {} Curtis-Holmes, J.%
\end{APACrefauthors}%
\unskip\
\newblock
\APACrefYearMonthDay{2005}{}{}.
\newblock
{\BBOQ}\APACrefatitle {Rapid responding increases belief bias: Evidence for the
  dual-process theory of reasoning} {Rapid responding increases belief bias:
  Evidence for the dual-process theory of reasoning}.{\BBCQ}
\newblock
\APACjournalVolNumPages{Thinking \& Reasoning}{11}{4}{382--389}.
\PrintBackRefs{\CurrentBib}

\bibitem [\protect \citeauthoryear {%
Evans%
, Handley%
, Over%
\BCBL {}\ \BBA {} Perham%
}{%
Evans%
\ \protect \BOthers {.}}{%
{\protect \APACyear {2002}}%
}]{%
evans2002background}
\APACinsertmetastar {%
evans2002background}%
\begin{APACrefauthors}%
Evans, J\BPBI S\BPBI B.%
, Handley, S\BPBI J.%
, Over, D\BPBI E.%
\BCBL {}\ \BBA {} Perham, N.%
\end{APACrefauthors}%
\unskip\
\newblock
\APACrefYearMonthDay{2002}{}{}.
\newblock
{\BBOQ}\APACrefatitle {Background beliefs in Bayesian inference} {Background
  beliefs in bayesian inference}.{\BBCQ}
\newblock
\APACjournalVolNumPages{Memory \& Cognition}{30}{2}{179--190}.
\PrintBackRefs{\CurrentBib}

\bibitem [\protect \citeauthoryear {%
Evans%
\ \BBA {} Perry%
}{%
Evans%
\ \BBA {} Perry%
}{%
{\protect \APACyear {1995}}%
}]{%
evans1995belief}
\APACinsertmetastar {%
evans1995belief}%
\begin{APACrefauthors}%
Evans, J\BPBI S\BPBI B.%
\BCBT {}\ \BBA {} Perry, T\BPBI S.%
\end{APACrefauthors}%
\unskip\
\newblock
\APACrefYearMonthDay{1995}{}{}.
\newblock
{\BBOQ}\APACrefatitle {Belief bias in children's reasoning.} {Belief bias in
  children's reasoning.}{\BBCQ}
\newblock
\APACjournalVolNumPages{Cahiers de Psychologie Cognitive/Current Psychology of
  Cognition}{}{}{}.
\PrintBackRefs{\CurrentBib}

\bibitem [\protect \citeauthoryear {%
Fodor%
\ \BBA {} Pylyshyn%
}{%
Fodor%
\ \BBA {} Pylyshyn%
}{%
{\protect \APACyear {1988}}%
}]{%
fodor88}
\APACinsertmetastar {%
fodor88}%
\begin{APACrefauthors}%
Fodor, J\BPBI A.%
\BCBT {}\ \BBA {} Pylyshyn, Z\BPBI W.%
\end{APACrefauthors}%
\unskip\
\newblock
\APACrefYearMonthDay{1988}{}{}.
\newblock
{\BBOQ}\APACrefatitle {Connectionism and cognitive architecture: A critical
  analysis} {Connectionism and cognitive architecture: A critical
  analysis}.{\BBCQ}
\newblock
\APACjournalVolNumPages{Cognition}{28}{}{3--71}.
\PrintBackRefs{\CurrentBib}

\bibitem [\protect \citeauthoryear {%
Gandhi%
\ \BBA {} Lake%
}{%
Gandhi%
\ \BBA {} Lake%
}{%
{\protect \APACyear {2019}}%
}]{%
gandhi2019mutual}
\APACinsertmetastar {%
gandhi2019mutual}%
\begin{APACrefauthors}%
Gandhi, K.%
\BCBT {}\ \BBA {} Lake, B\BPBI M.%
\end{APACrefauthors}%
\unskip\
\newblock
\APACrefYearMonthDay{2019}{}{}.
\newblock
{\BBOQ}\APACrefatitle {Mutual exclusivity as a challenge for neural networks}
  {Mutual exclusivity as a challenge for neural networks}.{\BBCQ}
\newblock
\APACjournalVolNumPages{arXiv preprint arXiv:1906.10197}{}{}{}.
\PrintBackRefs{\CurrentBib}

\bibitem [\protect \citeauthoryear {%
S.~Gershman%
\ \BBA {} Tenenbaum%
}{%
S.~Gershman%
\ \BBA {} Tenenbaum%
}{%
{\protect \APACyear {2015}}%
}]{%
gershman15}
\APACinsertmetastar {%
gershman15}%
\begin{APACrefauthors}%
Gershman, S.%
\BCBT {}\ \BBA {} Tenenbaum, J\BPBI B.%
\end{APACrefauthors}%
\unskip\
\newblock
\APACrefYearMonthDay{2015}{}{}.
\newblock
{\BBOQ}\APACrefatitle {Phrase similarity in humans and machines} {Phrase
  similarity in humans and machines}.{\BBCQ}
\newblock
\BIn{} \APACrefbtitle {Proceedings of the 37th Annual Conference of the
  Cognitive Science Society.} {Proceedings of the 37th annual conference of the
  cognitive science society.}
\PrintBackRefs{\CurrentBib}

\bibitem [\protect \citeauthoryear {%
S\BPBI J.~Gershman%
}{%
S\BPBI J.~Gershman%
}{%
{\protect \APACyear {2019}}%
}]{%
gershman2019generative}
\APACinsertmetastar {%
gershman2019generative}%
\begin{APACrefauthors}%
Gershman, S\BPBI J.%
\end{APACrefauthors}%
\unskip\
\newblock
\APACrefYearMonthDay{2019}{}{}.
\newblock
{\BBOQ}\APACrefatitle {The generative adversarial brain} {The generative
  adversarial brain}.{\BBCQ}
\newblock

\PrintBackRefs{\CurrentBib}

\bibitem [\protect \citeauthoryear {%
Gigerenzer%
\ \BBA {} Todd%
}{%
Gigerenzer%
\ \BBA {} Todd%
}{%
{\protect \APACyear {1999}}%
}]{%
gigerenzer1999simple}
\APACinsertmetastar {%
gigerenzer1999simple}%
\begin{APACrefauthors}%
Gigerenzer, G.%
\BCBT {}\ \BBA {} Todd, P\BPBI M.%
\end{APACrefauthors}%
\unskip\
\newblock
\APACrefYear{1999}.
\newblock
\APACrefbtitle {Simple heuristics that make us smart} {Simple heuristics that
  make us smart}.
\newblock
\APACaddressPublisher{}{Oxford University Press, USA}.
\PrintBackRefs{\CurrentBib}

\bibitem [\protect \citeauthoryear {%
Glockner%
, Shwartz%
\BCBL {}\ \BBA {} Goldberg%
}{%
Glockner%
\ \protect \BOthers {.}}{%
{\protect \APACyear {2018}}%
}]{%
glockner2018breaking}
\APACinsertmetastar {%
glockner2018breaking}%
\begin{APACrefauthors}%
Glockner, M.%
, Shwartz, V.%
\BCBL {}\ \BBA {} Goldberg, Y.%
\end{APACrefauthors}%
\unskip\
\newblock
\APACrefYearMonthDay{2018}{}{}.
\newblock
{\BBOQ}\APACrefatitle {Breaking nli systems with sentences that require simple
  lexical inferences} {Breaking nli systems with sentences that require simple
  lexical inferences}.{\BBCQ}
\newblock
\APACjournalVolNumPages{arXiv preprint arXiv:1805.02266}{}{}{}.
\PrintBackRefs{\CurrentBib}

\bibitem [\protect \citeauthoryear {%
Goodfellow%
, Shlens%
\BCBL {}\ \BBA {} Szegedy%
}{%
Goodfellow%
\ \protect \BOthers {.}}{%
{\protect \APACyear {2014}}%
}]{%
goodfellow2014explaining}
\APACinsertmetastar {%
goodfellow2014explaining}%
\begin{APACrefauthors}%
Goodfellow, I\BPBI J.%
, Shlens, J.%
\BCBL {}\ \BBA {} Szegedy, C.%
\end{APACrefauthors}%
\unskip\
\newblock
\APACrefYearMonthDay{2014}{}{}.
\newblock
{\BBOQ}\APACrefatitle {Explaining and harnessing adversarial examples}
  {Explaining and harnessing adversarial examples}.{\BBCQ}
\newblock
\APACjournalVolNumPages{arXiv preprint arXiv:1412.6572}{}{}{}.
\PrintBackRefs{\CurrentBib}

\bibitem [\protect \citeauthoryear {%
Groves%
\ \BBA {} Thompson%
}{%
Groves%
\ \BBA {} Thompson%
}{%
{\protect \APACyear {1970}}%
}]{%
groves1970habituation}
\APACinsertmetastar {%
groves1970habituation}%
\begin{APACrefauthors}%
Groves, P\BPBI M.%
\BCBT {}\ \BBA {} Thompson, R\BPBI F.%
\end{APACrefauthors}%
\unskip\
\newblock
\APACrefYearMonthDay{1970}{}{}.
\newblock
{\BBOQ}\APACrefatitle {Habituation: a dual-process theory.} {Habituation: a
  dual-process theory.}{\BBCQ}
\newblock
\APACjournalVolNumPages{Psychological review}{77}{5}{419}.
\PrintBackRefs{\CurrentBib}

\bibitem [\protect \citeauthoryear {%
Hill%
, Cho%
\BCBL {}\ \BBA {} Korhonen%
}{%
Hill%
\ \protect \BOthers {.}}{%
{\protect \APACyear {2016}}%
}]{%
Hill:2016uu}
\APACinsertmetastar {%
Hill:2016uu}%
\begin{APACrefauthors}%
Hill, F.%
, Cho, K.%
\BCBL {}\ \BBA {} Korhonen, A.%
\end{APACrefauthors}%
\unskip\
\newblock
\APACrefYearMonthDay{2016}{{\APACmonth{02}}}{}.
\newblock
{\BBOQ}\APACrefatitle {{Learning Distributed Representations of Sentences from
  Unlabelled Data}} {{Learning Distributed Representations of Sentences from
  Unlabelled Data}}.{\BBCQ}
\newblock

\PrintBackRefs{\CurrentBib}

\bibitem [\protect \citeauthoryear {%
Hornik%
}{%
Hornik%
}{%
{\protect \APACyear {1991}}%
}]{%
hornik1991approximation}
\APACinsertmetastar {%
hornik1991approximation}%
\begin{APACrefauthors}%
Hornik, K.%
\end{APACrefauthors}%
\unskip\
\newblock
\APACrefYearMonthDay{1991}{}{}.
\newblock
{\BBOQ}\APACrefatitle {Approximation capabilities of multilayer feedforward
  networks} {Approximation capabilities of multilayer feedforward
  networks}.{\BBCQ}
\newblock
\APACjournalVolNumPages{Neural networks}{4}{2}{251--257}.
\PrintBackRefs{\CurrentBib}

\bibitem [\protect \citeauthoryear {%
Johnson%
\ \protect \BOthers {.}}{%
Johnson%
\ \protect \BOthers {.}}{%
{\protect \APACyear {2017}}%
}]{%
johnson2017clevr}
\APACinsertmetastar {%
johnson2017clevr}%
\begin{APACrefauthors}%
Johnson, J.%
, Hariharan, B.%
, van~der Maaten, L.%
, Fei-Fei, L.%
, Lawrence~Zitnick, C.%
\BCBL {}\ \BBA {} Girshick, R.%
\end{APACrefauthors}%
\unskip\
\newblock
\APACrefYearMonthDay{2017}{}{}.
\newblock
{\BBOQ}\APACrefatitle {Clevr: A diagnostic dataset for compositional language
  and elementary visual reasoning} {Clevr: A diagnostic dataset for
  compositional language and elementary visual reasoning}.{\BBCQ}
\newblock
\BIn{} \APACrefbtitle {Proceedings of the IEEE Conference on Computer Vision
  and Pattern Recognition} {Proceedings of the ieee conference on computer
  vision and pattern recognition}\ (\BPGS\ 2901--2910).
\PrintBackRefs{\CurrentBib}

\bibitem [\protect \citeauthoryear {%
Johnson-Laird%
\ \BBA {} Steedman%
}{%
Johnson-Laird%
\ \BBA {} Steedman%
}{%
{\protect \APACyear {1978}}%
}]{%
johnson1978psychology}
\APACinsertmetastar {%
johnson1978psychology}%
\begin{APACrefauthors}%
Johnson-Laird, P\BPBI N.%
\BCBT {}\ \BBA {} Steedman, M.%
\end{APACrefauthors}%
\unskip\
\newblock
\APACrefYearMonthDay{1978}{}{}.
\newblock
{\BBOQ}\APACrefatitle {The psychology of syllogisms} {The psychology of
  syllogisms}.{\BBCQ}
\newblock
\APACjournalVolNumPages{Cognitive psychology}{10}{1}{64--99}.
\PrintBackRefs{\CurrentBib}

\bibitem [\protect \citeauthoryear {%
Jones%
\ \BBA {} Mewhort%
}{%
Jones%
\ \BBA {} Mewhort%
}{%
{\protect \APACyear {2007}}%
}]{%
jones2007representing}
\APACinsertmetastar {%
jones2007representing}%
\begin{APACrefauthors}%
Jones, M\BPBI N.%
\BCBT {}\ \BBA {} Mewhort, D\BPBI J.%
\end{APACrefauthors}%
\unskip\
\newblock
\APACrefYearMonthDay{2007}{}{}.
\newblock
{\BBOQ}\APACrefatitle {Representing word meaning and order information in a
  composite holographic lexicon.} {Representing word meaning and order
  information in a composite holographic lexicon.}{\BBCQ}
\newblock
\APACjournalVolNumPages{Psychological review}{114}{1}{1}.
\PrintBackRefs{\CurrentBib}

\bibitem [\protect \citeauthoryear {%
K{\'a}d{\'a}r%
, Chrupa{\l}a%
\BCBL {}\ \BBA {} Alishahi%
}{%
K{\'a}d{\'a}r%
\ \protect \BOthers {.}}{%
{\protect \APACyear {2017}}%
}]{%
kadar2017representation}
\APACinsertmetastar {%
kadar2017representation}%
\begin{APACrefauthors}%
K{\'a}d{\'a}r, A.%
, Chrupa{\l}a, G.%
\BCBL {}\ \BBA {} Alishahi, A.%
\end{APACrefauthors}%
\unskip\
\newblock
\APACrefYearMonthDay{2017}{}{}.
\newblock
{\BBOQ}\APACrefatitle {Representation of linguistic form and function in
  recurrent neural networks} {Representation of linguistic form and function in
  recurrent neural networks}.{\BBCQ}
\newblock
\APACjournalVolNumPages{Computational Linguistics}{43}{4}{761--780}.
\PrintBackRefs{\CurrentBib}

\bibitem [\protect \citeauthoryear {%
Kahneman%
}{%
Kahneman%
}{%
{\protect \APACyear {2011}}%
}]{%
kahneman2011thinking}
\APACinsertmetastar {%
kahneman2011thinking}%
\begin{APACrefauthors}%
Kahneman, D.%
\end{APACrefauthors}%
\unskip\
\newblock
\APACrefYear{2011}.
\newblock
\APACrefbtitle {Thinking, fast and slow} {Thinking, fast and slow}.
\newblock
\APACaddressPublisher{}{Macmillan}.
\PrintBackRefs{\CurrentBib}

\bibitem [\protect \citeauthoryear {%
Kang%
, Khot%
, Sabharwal%
\BCBL {}\ \BBA {} Hovy%
}{%
Kang%
\ \protect \BOthers {.}}{%
{\protect \APACyear {2018}}%
}]{%
kang2018adventure}
\APACinsertmetastar {%
kang2018adventure}%
\begin{APACrefauthors}%
Kang, D.%
, Khot, T.%
, Sabharwal, A.%
\BCBL {}\ \BBA {} Hovy, E.%
\end{APACrefauthors}%
\unskip\
\newblock
\APACrefYearMonthDay{2018}{}{}.
\newblock
{\BBOQ}\APACrefatitle {Adventure: Adversarial training for textual entailment
  with knowledge-guided examples} {Adventure: Adversarial training for textual
  entailment with knowledge-guided examples}.{\BBCQ}
\newblock
\APACjournalVolNumPages{arXiv preprint arXiv:1805.04680}{}{}{}.
\PrintBackRefs{\CurrentBib}

\bibitem [\protect \citeauthoryear {%
Karpathy%
, Johnson%
\BCBL {}\ \BBA {} Fei-Fei%
}{%
Karpathy%
\ \protect \BOthers {.}}{%
{\protect \APACyear {2015}}%
}]{%
karpathy2015visualizing}
\APACinsertmetastar {%
karpathy2015visualizing}%
\begin{APACrefauthors}%
Karpathy, A.%
, Johnson, J.%
\BCBL {}\ \BBA {} Fei-Fei, L.%
\end{APACrefauthors}%
\unskip\
\newblock
\APACrefYearMonthDay{2015}{}{}.
\newblock
{\BBOQ}\APACrefatitle {Visualizing and understanding recurrent networks}
  {Visualizing and understanding recurrent networks}.{\BBCQ}
\newblock
\APACjournalVolNumPages{arXiv preprint arXiv:1506.02078}{}{}{}.
\PrintBackRefs{\CurrentBib}

\bibitem [\protect \citeauthoryear {%
Kiros%
\ \protect \BOthers {.}}{%
Kiros%
\ \protect \BOthers {.}}{%
{\protect \APACyear {2015}}%
}]{%
kiros2015skip}
\APACinsertmetastar {%
kiros2015skip}%
\begin{APACrefauthors}%
Kiros, R.%
, Zhu, Y.%
, Salakhutdinov, R\BPBI R.%
, Zemel, R.%
, Urtasun, R.%
, Torralba, A.%
\BCBL {}\ \BBA {} Fidler, S.%
\end{APACrefauthors}%
\unskip\
\newblock
\APACrefYearMonthDay{2015}{}{}.
\newblock
{\BBOQ}\APACrefatitle {Skip-thought vectors} {Skip-thought vectors}.{\BBCQ}
\newblock
\BIn{} \APACrefbtitle {Advances in neural information processing systems}
  {Advances in neural information processing systems}\ (\BPGS\ 3294--3302).
\PrintBackRefs{\CurrentBib}

\bibitem [\protect \citeauthoryear {%
Lake%
\ \BBA {} Baroni%
}{%
Lake%
\ \BBA {} Baroni%
}{%
{\protect \APACyear {2017}}%
}]{%
lake17}
\APACinsertmetastar {%
lake17}%
\begin{APACrefauthors}%
Lake, B\BPBI M.%
\BCBT {}\ \BBA {} Baroni, M.%
\end{APACrefauthors}%
\unskip\
\newblock
\APACrefYearMonthDay{2017}{}{}.
\newblock
{\BBOQ}\APACrefatitle {Still not systematic after all these years: On the
  compositional skills of sequence-to-sequence recurrent networks} {Still not
  systematic after all these years: On the compositional skills of
  sequence-to-sequence recurrent networks}.{\BBCQ}
\newblock
\APACjournalVolNumPages{arXiv preprint arXiv:1711.00350}{}{}{}.
\PrintBackRefs{\CurrentBib}

\bibitem [\protect \citeauthoryear {%
Lake%
, Linzen%
\BCBL {}\ \BBA {} Baroni%
}{%
Lake%
\ \protect \BOthers {.}}{%
{\protect \APACyear {2019}}%
}]{%
lake2019human}
\APACinsertmetastar {%
lake2019human}%
\begin{APACrefauthors}%
Lake, B\BPBI M.%
, Linzen, T.%
\BCBL {}\ \BBA {} Baroni, M.%
\end{APACrefauthors}%
\unskip\
\newblock
\APACrefYearMonthDay{2019}{}{}.
\newblock
{\BBOQ}\APACrefatitle {Human few-shot learning of compositional instructions}
  {Human few-shot learning of compositional instructions}.{\BBCQ}
\newblock
\APACjournalVolNumPages{arXiv preprint arXiv:1901.04587}{}{}{}.
\PrintBackRefs{\CurrentBib}

\bibitem [\protect \citeauthoryear {%
Lake%
, Ullman%
, Tenenbaum%
\BCBL {}\ \BBA {} Gershman%
}{%
Lake%
\ \protect \BOthers {.}}{%
{\protect \APACyear {2018}}%
}]{%
lake18}
\APACinsertmetastar {%
lake18}%
\begin{APACrefauthors}%
Lake, B\BPBI M.%
, Ullman, T\BPBI D.%
, Tenenbaum, J\BPBI B.%
\BCBL {}\ \BBA {} Gershman, S\BPBI J.%
\end{APACrefauthors}%
\unskip\
\newblock
\APACrefYearMonthDay{2018}{}{}.
\newblock
{\BBOQ}\APACrefatitle {Building machines that learn and think like people}
  {Building machines that learn and think like people}.{\BBCQ}
\newblock
\APACjournalVolNumPages{Behavioral and Brain Sciences}{40}{}{}.
\PrintBackRefs{\CurrentBib}

\bibitem [\protect \citeauthoryear {%
Landauer%
\ \BBA {} Dumais%
}{%
Landauer%
\ \BBA {} Dumais%
}{%
{\protect \APACyear {1997}}%
}]{%
landauer1997solution}
\APACinsertmetastar {%
landauer1997solution}%
\begin{APACrefauthors}%
Landauer, T\BPBI K.%
\BCBT {}\ \BBA {} Dumais, S\BPBI T.%
\end{APACrefauthors}%
\unskip\
\newblock
\APACrefYearMonthDay{1997}{}{}.
\newblock
{\BBOQ}\APACrefatitle {A solution to Plato's problem: The latent semantic
  analysis theory of acquisition, induction, and representation of knowledge.}
  {A solution to plato's problem: The latent semantic analysis theory of
  acquisition, induction, and representation of knowledge.}{\BBCQ}
\newblock
\APACjournalVolNumPages{Psychological review}{104}{2}{211}.
\PrintBackRefs{\CurrentBib}

\bibitem [\protect \citeauthoryear {%
LeCun%
, Bengio%
\BCBL {}\ \BBA {} Hinton%
}{%
LeCun%
\ \protect \BOthers {.}}{%
{\protect \APACyear {2015}}%
}]{%
lecun2015deep}
\APACinsertmetastar {%
lecun2015deep}%
\begin{APACrefauthors}%
LeCun, Y.%
, Bengio, Y.%
\BCBL {}\ \BBA {} Hinton, G.%
\end{APACrefauthors}%
\unskip\
\newblock
\APACrefYearMonthDay{2015}{}{}.
\newblock
{\BBOQ}\APACrefatitle {Deep learning} {Deep learning}.{\BBCQ}
\newblock
\APACjournalVolNumPages{nature}{521}{7553}{436}.
\PrintBackRefs{\CurrentBib}

\bibitem [\protect \citeauthoryear {%
Li%
, Chen%
, Hovy%
\BCBL {}\ \BBA {} Jurafsky%
}{%
Li%
\ \protect \BOthers {.}}{%
{\protect \APACyear {2015}}%
}]{%
li2015visualizing}
\APACinsertmetastar {%
li2015visualizing}%
\begin{APACrefauthors}%
Li, J.%
, Chen, X.%
, Hovy, E.%
\BCBL {}\ \BBA {} Jurafsky, D.%
\end{APACrefauthors}%
\unskip\
\newblock
\APACrefYearMonthDay{2015}{}{}.
\newblock
{\BBOQ}\APACrefatitle {Visualizing and understanding neural models in nlp}
  {Visualizing and understanding neural models in nlp}.{\BBCQ}
\newblock
\APACjournalVolNumPages{arXiv preprint arXiv:1506.01066}{}{}{}.
\PrintBackRefs{\CurrentBib}

\bibitem [\protect \citeauthoryear {%
Lightfoot%
\ \BBA {} Julia%
}{%
Lightfoot%
\ \BBA {} Julia%
}{%
{\protect \APACyear {1984}}%
}]{%
lightfoot1984language}
\APACinsertmetastar {%
lightfoot1984language}%
\begin{APACrefauthors}%
Lightfoot, D.%
\BCBT {}\ \BBA {} Julia, P.%
\end{APACrefauthors}%
\unskip\
\newblock
\APACrefYearMonthDay{1984}{}{}.
\newblock
{\BBOQ}\APACrefatitle {The language lottery: Toward a biology of grammars} {The
  language lottery: Toward a biology of grammars}.{\BBCQ}
\newblock

\PrintBackRefs{\CurrentBib}

\bibitem [\protect \citeauthoryear {%
Linzen%
}{%
Linzen%
}{%
{\protect \APACyear {2019}}%
}]{%
linzen2019can}
\APACinsertmetastar {%
linzen2019can}%
\begin{APACrefauthors}%
Linzen, T.%
\end{APACrefauthors}%
\unskip\
\newblock
\APACrefYearMonthDay{2019}{}{}.
\newblock
{\BBOQ}\APACrefatitle {What can linguistics and deep learning contribute to
  each other? Response to Pater} {What can linguistics and deep learning
  contribute to each other? response to pater}.{\BBCQ}
\newblock
\APACjournalVolNumPages{Language}{}{}{}.
\PrintBackRefs{\CurrentBib}

\bibitem [\protect \citeauthoryear {%
Linzen%
, Dupoux%
\BCBL {}\ \BBA {} Goldberg%
}{%
Linzen%
\ \protect \BOthers {.}}{%
{\protect \APACyear {2016}}%
}]{%
linzen2016assessing}
\APACinsertmetastar {%
linzen2016assessing}%
\begin{APACrefauthors}%
Linzen, T.%
, Dupoux, E.%
\BCBL {}\ \BBA {} Goldberg, Y.%
\end{APACrefauthors}%
\unskip\
\newblock
\APACrefYearMonthDay{2016}{}{}.
\newblock
{\BBOQ}\APACrefatitle {Assessing the ability of LSTMs to learn syntax-sensitive
  dependencies} {Assessing the ability of lstms to learn syntax-sensitive
  dependencies}.{\BBCQ}
\newblock
\APACjournalVolNumPages{Transactions of the Association for Computational
  Linguistics}{4}{}{521--535}.
\PrintBackRefs{\CurrentBib}

\bibitem [\protect \citeauthoryear {%
Marcus%
}{%
Marcus%
}{%
{\protect \APACyear {2018}}%
}]{%
marcus2018deep}
\APACinsertmetastar {%
marcus2018deep}%
\begin{APACrefauthors}%
Marcus, G.%
\end{APACrefauthors}%
\unskip\
\newblock
\APACrefYearMonthDay{2018}{}{}.
\newblock
{\BBOQ}\APACrefatitle {Deep learning: A critical appraisal} {Deep learning: A
  critical appraisal}.{\BBCQ}
\newblock
\APACjournalVolNumPages{arXiv preprint arXiv:1801.00631}{}{}{}.
\PrintBackRefs{\CurrentBib}

\bibitem [\protect \citeauthoryear {%
Marelli%
\ \protect \BOthers {.}}{%
Marelli%
\ \protect \BOthers {.}}{%
{\protect \APACyear {2014}}%
}]{%
marelli2014semeval}
\APACinsertmetastar {%
marelli2014semeval}%
\begin{APACrefauthors}%
Marelli, M.%
, Bentivogli, L.%
, Baroni, M.%
, Bernardi, R.%
, Menini, S.%
\BCBL {}\ \BBA {} Zamparelli, R.%
\end{APACrefauthors}%
\unskip\
\newblock
\APACrefYearMonthDay{2014}{}{}.
\newblock
{\BBOQ}\APACrefatitle {Semeval-2014 task 1: Evaluation of compositional
  distributional semantic models on full sentences through semantic relatedness
  and textual entailment} {Semeval-2014 task 1: Evaluation of compositional
  distributional semantic models on full sentences through semantic relatedness
  and textual entailment}.{\BBCQ}
\newblock
\BIn{} \APACrefbtitle {Proceedings of the 8th international workshop on
  semantic evaluation (SemEval 2014)} {Proceedings of the 8th international
  workshop on semantic evaluation (semeval 2014)}\ (\BPGS\ 1--8).
\PrintBackRefs{\CurrentBib}

\bibitem [\protect \citeauthoryear {%
Martignon%
\ \BBA {} Hoffrage%
}{%
Martignon%
\ \BBA {} Hoffrage%
}{%
{\protect \APACyear {2002}}%
}]{%
martignon2002fast}
\APACinsertmetastar {%
martignon2002fast}%
\begin{APACrefauthors}%
Martignon, L.%
\BCBT {}\ \BBA {} Hoffrage, U.%
\end{APACrefauthors}%
\unskip\
\newblock
\APACrefYearMonthDay{2002}{}{}.
\newblock
{\BBOQ}\APACrefatitle {Fast, frugal, and fit: Simple heuristics for paired
  comparison} {Fast, frugal, and fit: Simple heuristics for paired
  comparison}.{\BBCQ}
\newblock
\APACjournalVolNumPages{Theory and Decision}{52}{1}{29--71}.
\PrintBackRefs{\CurrentBib}

\bibitem [\protect \citeauthoryear {%
McCoy%
, Pavlick%
\BCBL {}\ \BBA {} Linzen%
}{%
McCoy%
\ \protect \BOthers {.}}{%
{\protect \APACyear {2019}}%
}]{%
mccoy2019right}
\APACinsertmetastar {%
mccoy2019right}%
\begin{APACrefauthors}%
McCoy, R\BPBI T.%
, Pavlick, E.%
\BCBL {}\ \BBA {} Linzen, T.%
\end{APACrefauthors}%
\unskip\
\newblock
\APACrefYearMonthDay{2019}{}{}.
\newblock
{\BBOQ}\APACrefatitle {Right for the Wrong Reasons: Diagnosing Syntactic
  Heuristics in Natural Language Inference} {Right for the wrong reasons:
  Diagnosing syntactic heuristics in natural language inference}.{\BBCQ}
\newblock
\APACjournalVolNumPages{arXiv preprint arXiv:1902.01007}{}{}{}.
\PrintBackRefs{\CurrentBib}

\bibitem [\protect \citeauthoryear {%
Mikolov%
, Sutskever%
, Chen%
, Corrado%
\BCBL {}\ \BBA {} Dean%
}{%
Mikolov%
\ \protect \BOthers {.}}{%
{\protect \APACyear {2013}}%
}]{%
mikolov2013distributed}
\APACinsertmetastar {%
mikolov2013distributed}%
\begin{APACrefauthors}%
Mikolov, T.%
, Sutskever, I.%
, Chen, K.%
, Corrado, G\BPBI S.%
\BCBL {}\ \BBA {} Dean, J.%
\end{APACrefauthors}%
\unskip\
\newblock
\APACrefYearMonthDay{2013}{}{}.
\newblock
{\BBOQ}\APACrefatitle {Distributed representations of words and phrases and
  their compositionality} {Distributed representations of words and phrases and
  their compositionality}.{\BBCQ}
\newblock
\BIn{} \APACrefbtitle {Advances in neural information processing systems}
  {Advances in neural information processing systems}\ (\BPGS\ 3111--3119).
\PrintBackRefs{\CurrentBib}

\bibitem [\protect \citeauthoryear {%
Mitchell%
}{%
Mitchell%
}{%
{\protect \APACyear {1980}}%
}]{%
mitchell1980need}
\APACinsertmetastar {%
mitchell1980need}%
\begin{APACrefauthors}%
Mitchell, T\BPBI M.%
\end{APACrefauthors}%
\unskip\
\newblock
\APACrefYear{1980}.
\newblock
\APACrefbtitle {The need for biases in learning generalizations} {The need for
  biases in learning generalizations}.
\newblock
\APACaddressPublisher{}{Department of Computer Science, Laboratory for Computer
  Science Research~…}.
\PrintBackRefs{\CurrentBib}

\bibitem [\protect \citeauthoryear {%
Mnih%
\ \protect \BOthers {.}}{%
Mnih%
\ \protect \BOthers {.}}{%
{\protect \APACyear {2015}}%
}]{%
mnih2015human}
\APACinsertmetastar {%
mnih2015human}%
\begin{APACrefauthors}%
Mnih, V.%
, Kavukcuoglu, K.%
, Silver, D.%
, Rusu, A\BPBI A.%
, Veness, J.%
, Bellemare, M\BPBI G.%
\BDBL {}others%
\end{APACrefauthors}%
\unskip\
\newblock
\APACrefYearMonthDay{2015}{}{}.
\newblock
{\BBOQ}\APACrefatitle {Human-level control through deep reinforcement learning}
  {Human-level control through deep reinforcement learning}.{\BBCQ}
\newblock
\APACjournalVolNumPages{Nature}{518}{7540}{529}.
\PrintBackRefs{\CurrentBib}

\bibitem [\protect \citeauthoryear {%
Nie%
, Wang%
\BCBL {}\ \BBA {} Bansal%
}{%
Nie%
\ \protect \BOthers {.}}{%
{\protect \APACyear {2019}}%
}]{%
nie2019analyzing}
\APACinsertmetastar {%
nie2019analyzing}%
\begin{APACrefauthors}%
Nie, Y.%
, Wang, Y.%
\BCBL {}\ \BBA {} Bansal, M.%
\end{APACrefauthors}%
\unskip\
\newblock
\APACrefYearMonthDay{2019}{}{}.
\newblock
{\BBOQ}\APACrefatitle {Analyzing compositionality-sensitivity of nli models}
  {Analyzing compositionality-sensitivity of nli models}.{\BBCQ}
\newblock
\BIn{} \APACrefbtitle {Proceedings of the AAAI Conference on Artificial
  Intelligence} {Proceedings of the aaai conference on artificial
  intelligence}\ (\BVOL~33, \BPGS\ 6867--6874).
\PrintBackRefs{\CurrentBib}

\bibitem [\protect \citeauthoryear {%
Ommer%
\ \BBA {} Buhmann%
}{%
Ommer%
\ \BBA {} Buhmann%
}{%
{\protect \APACyear {2009}}%
}]{%
ommer2009learning}
\APACinsertmetastar {%
ommer2009learning}%
\begin{APACrefauthors}%
Ommer, B.%
\BCBT {}\ \BBA {} Buhmann, J.%
\end{APACrefauthors}%
\unskip\
\newblock
\APACrefYearMonthDay{2009}{}{}.
\newblock
{\BBOQ}\APACrefatitle {Learning the compositional nature of visual object
  categories for recognition} {Learning the compositional nature of visual
  object categories for recognition}.{\BBCQ}
\newblock
\APACjournalVolNumPages{IEEE Transactions on Pattern Analysis and Machine
  Intelligence}{32}{3}{501--516}.
\PrintBackRefs{\CurrentBib}

\bibitem [\protect \citeauthoryear {%
Oord%
\ \protect \BOthers {.}}{%
Oord%
\ \protect \BOthers {.}}{%
{\protect \APACyear {2016}}%
}]{%
oord2016wavenet}
\APACinsertmetastar {%
oord2016wavenet}%
\begin{APACrefauthors}%
Oord, A\BPBI v\BPBI d.%
, Dieleman, S.%
, Zen, H.%
, Simonyan, K.%
, Vinyals, O.%
, Graves, A.%
\BDBL {}Kavukcuoglu, K.%
\end{APACrefauthors}%
\unskip\
\newblock
\APACrefYearMonthDay{2016}{}{}.
\newblock
{\BBOQ}\APACrefatitle {Wavenet: A generative model for raw audio} {Wavenet: A
  generative model for raw audio}.{\BBCQ}
\newblock
\APACjournalVolNumPages{arXiv preprint arXiv:1609.03499}{}{}{}.
\PrintBackRefs{\CurrentBib}

\bibitem [\protect \citeauthoryear {%
Pearl%
\ \BBA {} Goldwater%
}{%
Pearl%
\ \BBA {} Goldwater%
}{%
{\protect \APACyear {2016}}%
}]{%
pearl2016statistical}
\APACinsertmetastar {%
pearl2016statistical}%
\begin{APACrefauthors}%
Pearl, L.%
\BCBT {}\ \BBA {} Goldwater, S.%
\end{APACrefauthors}%
\unskip\
\newblock
\APACrefYearMonthDay{2016}{}{}.
\newblock
{\BBOQ}\APACrefatitle {Statistical learning, inductive bias, and Bayesian
  inference in language acquisition} {Statistical learning, inductive bias, and
  bayesian inference in language acquisition}.{\BBCQ}
\newblock

\PrintBackRefs{\CurrentBib}

\bibitem [\protect \citeauthoryear {%
Pennington%
, Socher%
\BCBL {}\ \BBA {} Manning%
}{%
Pennington%
\ \protect \BOthers {.}}{%
{\protect \APACyear {2014}}%
}]{%
pennington14}
\APACinsertmetastar {%
pennington14}%
\begin{APACrefauthors}%
Pennington, J.%
, Socher, R.%
\BCBL {}\ \BBA {} Manning, C.%
\end{APACrefauthors}%
\unskip\
\newblock
\APACrefYearMonthDay{2014}{}{}.
\newblock
{\BBOQ}\APACrefatitle {Glove: Global vectors for word representation} {Glove:
  Global vectors for word representation}.{\BBCQ}
\newblock
\BIn{} \APACrefbtitle {Proceedings of the 2014 conference on empirical methods
  in natural language processing (EMNLP)} {Proceedings of the 2014 conference
  on empirical methods in natural language processing (emnlp)}\ (\BPGS\
  1532--1543).
\PrintBackRefs{\CurrentBib}

\bibitem [\protect \citeauthoryear {%
Pereira%
, Gershman%
, Ritter%
\BCBL {}\ \BBA {} Botvinick%
}{%
Pereira%
\ \protect \BOthers {.}}{%
{\protect \APACyear {2016}}%
}]{%
pereira2016comparative}
\APACinsertmetastar {%
pereira2016comparative}%
\begin{APACrefauthors}%
Pereira, F.%
, Gershman, S.%
, Ritter, S.%
\BCBL {}\ \BBA {} Botvinick, M.%
\end{APACrefauthors}%
\unskip\
\newblock
\APACrefYearMonthDay{2016}{}{}.
\newblock
{\BBOQ}\APACrefatitle {A comparative evaluation of off-the-shelf distributed
  semantic representations for modelling behavioural data} {A comparative
  evaluation of off-the-shelf distributed semantic representations for
  modelling behavioural data}.{\BBCQ}
\newblock
\APACjournalVolNumPages{Cognitive neuropsychology}{33}{3-4}{175--190}.
\PrintBackRefs{\CurrentBib}

\bibitem [\protect \citeauthoryear {%
Redington%
, Crater%
\BCBL {}\ \BBA {} Finch%
}{%
Redington%
\ \protect \BOthers {.}}{%
{\protect \APACyear {1998}}%
}]{%
redington1998distributional}
\APACinsertmetastar {%
redington1998distributional}%
\begin{APACrefauthors}%
Redington, M.%
, Crater, N.%
\BCBL {}\ \BBA {} Finch, S.%
\end{APACrefauthors}%
\unskip\
\newblock
\APACrefYearMonthDay{1998}{}{}.
\newblock
{\BBOQ}\APACrefatitle {Distributional information: A powerful cue for acquiring
  syntactic categories} {Distributional information: A powerful cue for
  acquiring syntactic categories}.{\BBCQ}
\newblock
\APACjournalVolNumPages{Cognitive science}{22}{4}{425--469}.
\PrintBackRefs{\CurrentBib}

\bibitem [\protect \citeauthoryear {%
Ritter%
, Barrett%
, Santoro%
\BCBL {}\ \BBA {} Botvinick%
}{%
Ritter%
\ \protect \BOthers {.}}{%
{\protect \APACyear {2017}}%
}]{%
ritter17}
\APACinsertmetastar {%
ritter17}%
\begin{APACrefauthors}%
Ritter, S.%
, Barrett, D\BPBI G.%
, Santoro, A.%
\BCBL {}\ \BBA {} Botvinick, M\BPBI M.%
\end{APACrefauthors}%
\unskip\
\newblock
\APACrefYearMonthDay{2017}{}{}.
\newblock
{\BBOQ}\APACrefatitle {Cognitive Psychology for Deep Neural Networks: A Shape
  Bias Case Study} {Cognitive psychology for deep neural networks: A shape bias
  case study}.{\BBCQ}
\newblock
\BIn{} \APACrefbtitle {International Conference on Machine Learning}
  {International conference on machine learning}\ (\BPGS\ 2940--2949).
\PrintBackRefs{\CurrentBib}

\bibitem [\protect \citeauthoryear {%
Ruder%
}{%
Ruder%
}{%
{\protect \APACyear {2016}}%
}]{%
ruder2016overview}
\APACinsertmetastar {%
ruder2016overview}%
\begin{APACrefauthors}%
Ruder, S.%
\end{APACrefauthors}%
\unskip\
\newblock
\APACrefYearMonthDay{2016}{}{}.
\newblock
{\BBOQ}\APACrefatitle {An overview of gradient descent optimization algorithms}
  {An overview of gradient descent optimization algorithms}.{\BBCQ}
\newblock
\APACjournalVolNumPages{arXiv preprint arXiv:1609.04747}{}{}{}.
\PrintBackRefs{\CurrentBib}

\bibitem [\protect \citeauthoryear {%
Rumelhart%
\ \BBA {} Abrahamson%
}{%
Rumelhart%
\ \BBA {} Abrahamson%
}{%
{\protect \APACyear {1973}}%
}]{%
rumelhart1973model}
\APACinsertmetastar {%
rumelhart1973model}%
\begin{APACrefauthors}%
Rumelhart, D\BPBI E.%
\BCBT {}\ \BBA {} Abrahamson, A\BPBI A.%
\end{APACrefauthors}%
\unskip\
\newblock
\APACrefYearMonthDay{1973}{}{}.
\newblock
{\BBOQ}\APACrefatitle {A model for analogical reasoning} {A model for
  analogical reasoning}.{\BBCQ}
\newblock
\APACjournalVolNumPages{Cognitive Psychology}{5}{1}{1--28}.
\PrintBackRefs{\CurrentBib}

\bibitem [\protect \citeauthoryear {%
Russakovsky%
\ \protect \BOthers {.}}{%
Russakovsky%
\ \protect \BOthers {.}}{%
{\protect \APACyear {2015}}%
}]{%
russakovsky2015imagenet}
\APACinsertmetastar {%
russakovsky2015imagenet}%
\begin{APACrefauthors}%
Russakovsky, O.%
, Deng, J.%
, Su, H.%
, Krause, J.%
, Satheesh, S.%
, Ma, S.%
\BDBL {}others%
\end{APACrefauthors}%
\unskip\
\newblock
\APACrefYearMonthDay{2015}{}{}.
\newblock
{\BBOQ}\APACrefatitle {Imagenet large scale visual recognition challenge}
  {Imagenet large scale visual recognition challenge}.{\BBCQ}
\newblock
\APACjournalVolNumPages{International journal of computer
  vision}{115}{3}{211--252}.
\PrintBackRefs{\CurrentBib}

\bibitem [\protect \citeauthoryear {%
Seidenberg%
}{%
Seidenberg%
}{%
{\protect \APACyear {1997}}%
}]{%
seidenberg1997language}
\APACinsertmetastar {%
seidenberg1997language}%
\begin{APACrefauthors}%
Seidenberg, M\BPBI S.%
\end{APACrefauthors}%
\unskip\
\newblock
\APACrefYearMonthDay{1997}{}{}.
\newblock
{\BBOQ}\APACrefatitle {Language acquisition and use: Learning and applying
  probabilistic constraints} {Language acquisition and use: Learning and
  applying probabilistic constraints}.{\BBCQ}
\newblock
\APACjournalVolNumPages{Science}{275}{5306}{1599--1603}.
\PrintBackRefs{\CurrentBib}

\bibitem [\protect \citeauthoryear {%
Siegelmann%
\ \BBA {} Sontag%
}{%
Siegelmann%
\ \BBA {} Sontag%
}{%
{\protect \APACyear {1995}}%
}]{%
siegelmann1995computational}
\APACinsertmetastar {%
siegelmann1995computational}%
\begin{APACrefauthors}%
Siegelmann, H\BPBI T.%
\BCBT {}\ \BBA {} Sontag, E\BPBI D.%
\end{APACrefauthors}%
\unskip\
\newblock
\APACrefYearMonthDay{1995}{}{}.
\newblock
{\BBOQ}\APACrefatitle {On the computational power of neural nets} {On the
  computational power of neural nets}.{\BBCQ}
\newblock
\APACjournalVolNumPages{Journal of computer and system
  sciences}{50}{1}{132--150}.
\PrintBackRefs{\CurrentBib}

\bibitem [\protect \citeauthoryear {%
Simon%
}{%
Simon%
}{%
{\protect \APACyear {1991}}%
}]{%
simon1991bounded}
\APACinsertmetastar {%
simon1991bounded}%
\begin{APACrefauthors}%
Simon, H\BPBI A.%
\end{APACrefauthors}%
\unskip\
\newblock
\APACrefYearMonthDay{1991}{}{}.
\newblock
{\BBOQ}\APACrefatitle {Bounded rationality and organizational learning}
  {Bounded rationality and organizational learning}.{\BBCQ}
\newblock
\APACjournalVolNumPages{Organization science}{2}{1}{125--134}.
\PrintBackRefs{\CurrentBib}

\bibitem [\protect \citeauthoryear {%
{\c{S}}im{\c{s}}ek%
}{%
{\c{S}}im{\c{s}}ek%
}{%
{\protect \APACyear {2013}}%
}]{%
csimcsek2013linear}
\APACinsertmetastar {%
csimcsek2013linear}%
\begin{APACrefauthors}%
{\c{S}}im{\c{s}}ek, {\"O}.%
\end{APACrefauthors}%
\unskip\
\newblock
\APACrefYearMonthDay{2013}{}{}.
\newblock
{\BBOQ}\APACrefatitle {Linear decision rule as aspiration for simple decision
  heuristics} {Linear decision rule as aspiration for simple decision
  heuristics}.{\BBCQ}
\newblock
\BIn{} \APACrefbtitle {Advances in neural information processing systems}
  {Advances in neural information processing systems}\ (\BPGS\ 2904--2912).
\PrintBackRefs{\CurrentBib}

\bibitem [\protect \citeauthoryear {%
{\c{S}}im{\c{s}}ek%
, Algorta%
\BCBL {}\ \BBA {} Kothiyal%
}{%
{\c{S}}im{\c{s}}ek%
\ \protect \BOthers {.}}{%
{\protect \APACyear {2016}}%
}]{%
csimcsek2016most}
\APACinsertmetastar {%
csimcsek2016most}%
\begin{APACrefauthors}%
{\c{S}}im{\c{s}}ek, {\"O}.%
, Algorta, S.%
\BCBL {}\ \BBA {} Kothiyal, A.%
\end{APACrefauthors}%
\unskip\
\newblock
\APACrefYearMonthDay{2016}{}{}.
\newblock
{\BBOQ}\APACrefatitle {Why most decisions are easy in tetris-and perhaps in
  other sequential decision problems, as well} {Why most decisions are easy in
  tetris-and perhaps in other sequential decision problems, as well}.{\BBCQ}
\newblock
\APACjournalVolNumPages{Proceedings of Machine Learning
  Research}{48}{}{1757--1765}.
\PrintBackRefs{\CurrentBib}

\bibitem [\protect \citeauthoryear {%
Singh%
}{%
Singh%
}{%
{\protect \APACyear {1992}}%
}]{%
singh1992transfer}
\APACinsertmetastar {%
singh1992transfer}%
\begin{APACrefauthors}%
Singh, S\BPBI P.%
\end{APACrefauthors}%
\unskip\
\newblock
\APACrefYearMonthDay{1992}{}{}.
\newblock
{\BBOQ}\APACrefatitle {Transfer of learning by composing solutions of elemental
  sequential tasks} {Transfer of learning by composing solutions of elemental
  sequential tasks}.{\BBCQ}
\newblock
\APACjournalVolNumPages{Machine Learning}{8}{3-4}{323--339}.
\PrintBackRefs{\CurrentBib}

\bibitem [\protect \citeauthoryear {%
Socher%
, Manning%
\BCBL {}\ \BBA {} Ng%
}{%
Socher%
\ \protect \BOthers {.}}{%
{\protect \APACyear {2010}}%
}]{%
socher2010learning}
\APACinsertmetastar {%
socher2010learning}%
\begin{APACrefauthors}%
Socher, R.%
, Manning, C\BPBI D.%
\BCBL {}\ \BBA {} Ng, A\BPBI Y.%
\end{APACrefauthors}%
\unskip\
\newblock
\APACrefYearMonthDay{2010}{}{}.
\newblock
{\BBOQ}\APACrefatitle {Learning continuous phrase representations and syntactic
  parsing with recursive neural networks} {Learning continuous phrase
  representations and syntactic parsing with recursive neural networks}.{\BBCQ}
\newblock
\BIn{} \APACrefbtitle {Proceedings of the NIPS-2010 Deep Learning and
  Unsupervised Feature Learning Workshop} {Proceedings of the nips-2010 deep
  learning and unsupervised feature learning workshop}\ (\BVOL\ 2010, \BPGS\
  1--9).
\PrintBackRefs{\CurrentBib}

\bibitem [\protect \citeauthoryear {%
Steyvers%
}{%
Steyvers%
}{%
{\protect \APACyear {2006}}%
}]{%
steyvers2006multidimensional}
\APACinsertmetastar {%
steyvers2006multidimensional}%
\begin{APACrefauthors}%
Steyvers, M.%
\end{APACrefauthors}%
\unskip\
\newblock
\APACrefYearMonthDay{2006}{}{}.
\newblock
{\BBOQ}\APACrefatitle {Multidimensional scaling} {Multidimensional
  scaling}.{\BBCQ}
\newblock
\APACjournalVolNumPages{Encyclopedia of cognitive science}{}{}{}.
\PrintBackRefs{\CurrentBib}

\bibitem [\protect \citeauthoryear {%
Todd%
\ \BBA {} Gigerenzer%
}{%
Todd%
\ \BBA {} Gigerenzer%
}{%
{\protect \APACyear {2007}}%
}]{%
todd2007environments}
\APACinsertmetastar {%
todd2007environments}%
\begin{APACrefauthors}%
Todd, P\BPBI M.%
\BCBT {}\ \BBA {} Gigerenzer, G.%
\end{APACrefauthors}%
\unskip\
\newblock
\APACrefYearMonthDay{2007}{}{}.
\newblock
{\BBOQ}\APACrefatitle {Environments that make us smart: Ecological rationality}
  {Environments that make us smart: Ecological rationality}.{\BBCQ}
\newblock
\APACjournalVolNumPages{Current directions in psychological
  science}{16}{3}{167--171}.
\PrintBackRefs{\CurrentBib}

\bibitem [\protect \citeauthoryear {%
White%
, Rastogi%
, Duh%
\BCBL {}\ \BBA {} Van~Durme%
}{%
White%
\ \protect \BOthers {.}}{%
{\protect \APACyear {2017}}%
}]{%
white2017inference}
\APACinsertmetastar {%
white2017inference}%
\begin{APACrefauthors}%
White, A\BPBI S.%
, Rastogi, P.%
, Duh, K.%
\BCBL {}\ \BBA {} Van~Durme, B.%
\end{APACrefauthors}%
\unskip\
\newblock
\APACrefYearMonthDay{2017}{}{}.
\newblock
{\BBOQ}\APACrefatitle {Inference is everything: Recasting semantic resources
  into a unified evaluation framework} {Inference is everything: Recasting
  semantic resources into a unified evaluation framework}.{\BBCQ}
\newblock
\BIn{} \APACrefbtitle {Proceedings of the Eighth International Joint Conference
  on Natural Language Processing (Volume 1: Long Papers)} {Proceedings of the
  eighth international joint conference on natural language processing (volume
  1: Long papers)}\ (\BPGS\ 996--1005).
\PrintBackRefs{\CurrentBib}

\bibitem [\protect \citeauthoryear {%
Yosinski%
, Clune%
, Nguyen%
, Fuchs%
\BCBL {}\ \BBA {} Lipson%
}{%
Yosinski%
\ \protect \BOthers {.}}{%
{\protect \APACyear {2015}}%
}]{%
yosinski2015understanding}
\APACinsertmetastar {%
yosinski2015understanding}%
\begin{APACrefauthors}%
Yosinski, J.%
, Clune, J.%
, Nguyen, A.%
, Fuchs, T.%
\BCBL {}\ \BBA {} Lipson, H.%
\end{APACrefauthors}%
\unskip\
\newblock
\APACrefYearMonthDay{2015}{}{}.
\newblock
{\BBOQ}\APACrefatitle {Understanding neural networks through deep
  visualization} {Understanding neural networks through deep
  visualization}.{\BBCQ}
\newblock
\APACjournalVolNumPages{arXiv preprint arXiv:1506.06579}{}{}{}.
\PrintBackRefs{\CurrentBib}

\bibitem [\protect \citeauthoryear {%
Zeiler%
\ \BBA {} Fergus%
}{%
Zeiler%
\ \BBA {} Fergus%
}{%
{\protect \APACyear {2014}}%
}]{%
zeiler2014visualizing}
\APACinsertmetastar {%
zeiler2014visualizing}%
\begin{APACrefauthors}%
Zeiler, M\BPBI D.%
\BCBT {}\ \BBA {} Fergus, R.%
\end{APACrefauthors}%
\unskip\
\newblock
\APACrefYearMonthDay{2014}{}{}.
\newblock
{\BBOQ}\APACrefatitle {Visualizing and understanding convolutional networks}
  {Visualizing and understanding convolutional networks}.{\BBCQ}
\newblock
\BIn{} \APACrefbtitle {European conference on computer vision} {European
  conference on computer vision}\ (\BPGS\ 818--833).
\PrintBackRefs{\CurrentBib}

\bibitem [\protect \citeauthoryear {%
Zhao%
, Dua%
\BCBL {}\ \BBA {} Singh%
}{%
Zhao%
\ \protect \BOthers {.}}{%
{\protect \APACyear {2017}}%
}]{%
zhao2017generating}
\APACinsertmetastar {%
zhao2017generating}%
\begin{APACrefauthors}%
Zhao, Z.%
, Dua, D.%
\BCBL {}\ \BBA {} Singh, S.%
\end{APACrefauthors}%
\unskip\
\newblock
\APACrefYearMonthDay{2017}{}{}.
\newblock
{\BBOQ}\APACrefatitle {Generating natural adversarial examples} {Generating
  natural adversarial examples}.{\BBCQ}
\newblock
\APACjournalVolNumPages{arXiv preprint arXiv:1710.11342}{}{}{}.
\PrintBackRefs{\CurrentBib}

\end{thebibliography}

\end{document}